\documentclass[10pt,twocolumn,letterpaper]{article}

\usepackage[pagenumbers]{cvpr} %

\definecolor{cvprblue}{rgb}{0.21,0.49,0.74}
\usepackage[pagebackref=true,breaklinks,colorlinks,allcolors=cvprblue]{hyperref}
\usepackage[accsupp]{axessibility}  %

\def\confName{CVPR}
\def\confYear{2025}

\title{Generative Photomontage}

\author{Sean J. Liu\textsuperscript{1}
\qquad
Nupur Kumari\textsuperscript{1}
\qquad
Ariel Shamir\textsuperscript{2}
\qquad
Jun-Yan Zhu\textsuperscript{1}\\\\
\textsuperscript{1}Carnegie Mellon University
\qquad
\textsuperscript{2}Reichman University
}

\usepackage{color,xcolor}
\usepackage{epsfig}
\usepackage{graphicx}

\usepackage{microtype}
\usepackage[font=small]{caption}
\usepackage{arydshln}
\usepackage{tabularx}
\usepackage{adjustbox}
\usepackage{array}
\usepackage{booktabs}
\usepackage{colortbl}
\usepackage{float,wrapfig}
\usepackage{hhline}
\usepackage{multirow}
\usepackage{subcaption} %
\usepackage[percent]{overpic}

\usepackage{breqn}

\usepackage{duckuments}
\usepackage{amsmath}

\usepackage{bm}
\usepackage{nicefrac}
\usepackage{microtype}
\usepackage{dsfont}
\usepackage{changepage}
\usepackage{extramarks}
\usepackage{fancyhdr}
\usepackage{setspace}
\usepackage{soul}
\usepackage{xspace}
\usepackage{hhline}
\usepackage{algorithmicx}
\usepackage{algpseudocode}
\usepackage{pifont}
\usepackage{booktabs}
\usepackage{multirow}

\usepackage{makecell}

\usepackage{enumitem}
\usepackage[title]{appendix}

\newcommand{\myparagraph}[1]{\vspace{5pt} \noindent \textbf{#1} \ }

\newcommand{\shortcite}{\cite}

\begin{document}

\twocolumn[{%
\renewcommand\twocolumn[1][]{#1}%
\maketitle

\begin{center}
    \vspace{-20pt}
    \centering
    \includegraphics[width=\textwidth]{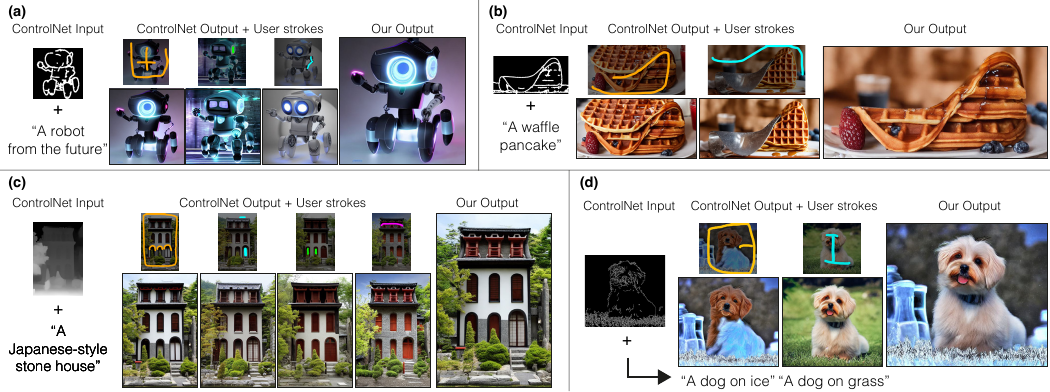}
    \captionof{figure}{We introduce Generative Photomontage, a framework that allows users to create their desired image by \emph{compositing} multiple generated images. 
      Given a stack of ControlNet-generated images using the same input condition and different seeds,  
     users select desired regions from different images within the stack. Our method takes in the user strokes, solves for a segmentation across the stack using diffusion features, and then composites them using a new feature-space blending method. 
     Our method offers users fine-grained control over the final image and enables various applications, such as generating unseen appearance combinations (a, c), correcting shapes and removing artifacts (b, d).} 
      \label{fig:teaser}
\end{center}
}]

\maketitle

\begin{abstract}
Text-to-image models are powerful tools for image creation. However, the generation process is akin to a dice roll and makes it difficult to achieve a \textit{single} image that captures everything a user wants. In this paper, we propose a framework for creating the desired image by \textit{compositing} it from various parts of generated images, in essence forming a \emph{Generative Photomontage}. Given a stack of images generated by ControlNet using the same input condition and different seeds, we let users select desired parts from the generated results using a brush stroke interface. 
We introduce a novel technique that takes in the user's brush strokes, segments the generated images using a graph-based optimization in diffusion feature space, and then composites the segmented regions via a new feature-space blending method. Our method faithfully preserves the user-selected regions while compositing them harmoniously. We demonstrate that our flexible framework can be used for many applications, including generating new appearance combinations, fixing incorrect shapes and artifacts, and improving prompt alignment. We show compelling results for each application and demonstrate that our method outperforms existing image blending methods and various baselines. 
\end{abstract}

\section{Introduction}
\label{sec:intro}
Text-to-image models~\cite{Rombach_2022_CVPR,zhang2023adding} can generate visually compelling images from simple input conditions, such as text prompts and sketches, making them a powerful tool for image synthesis and creative exploration. 

However, these models may not achieve exactly what a user envisions, due to the ambiguity in mapping from lower-dimensional input space (e.g., text, sketch) to high-dimensional pixel space.
For example, the prompt ``a robot from the future'' can map to any sample in a large space of robot images, that is usually sampled using different random seeds in the diffusion process. From the user's perspective, this procedure is akin to a dice roll.
In particular,
it is often challenging to achieve a \textit{single} image that includes everything the user wants: the user may like one part of the robot from one result and another part in a different result. They may also like the background in yet a third result. 

Many works add various conditions to text-to-image models for greater user control~\cite{zhang2023adding,mou2024t2i}, such as edges and depth maps. %
While these approaches restrict the output space to better match the additional user inputs,
the process is still akin to a dice roll
(albeit with a constrained die).
For example, using the same edge map and text prompt, ControlNet~\cite{zhang2023adding} can generate a range of outputs that differ in lighting, appearance, and backgrounds. Some results might contain desirable visual elements, while others could contain artifacts or fail to adhere closely to the input conditions.
While one can create numerous variations using different random seeds (i.e., re-roll the dice), 
such a trial-and-error process offers limited user control and makes it challenging to achieve a completely satisfactory result.

In this paper, we propose a 
different approach -- we 
suggest the possibility of synthesizing the desired image by \textit{compositing} it from different parts of generated images. 
We refer to the final result as a \emph{Generative Photomontage}, inspired by the seminal work of Interactive Digital Photomontage~\cite{agarwala2004interactive}. In our approach, users can first generate many results (roll the dice first) and then choose exactly what they want (composite across the dice rolls),
which gives users fine-grained control over the final output and significantly increases the likelihood of achieving their desired result.
Our key idea is to treat generated images as intermediate outputs, let users select desired parts from the generated results, 
and then composite the user-selected regions to form the final image.

Our framework begins with a \emph{stack} of images from ControlNet, generated by using 
the same input condition and different seeds,
and lets users choose parts they like from different images via simple brush strokes.
Our key insight is that these images share common spatial structures from the same input condition, which can be leveraged for composition.
We propose a novel technique that takes in the user's brush strokes, segments the image parts in diffusion feature space, and then composites these parts during a final denoising process. Specifically, given users' sparse scribbles, we formulate a multi-label graph-based optimization in diffusion feature space, grouping regions with similar diffusion features while satisfying user inputs. We then introduce a new feature injection and mixing method to composite the segmented regions.
Our method accurately preserves the user-selected regions while harmoniously blending them together.

The advantages of using our approach are two-fold. First is the user interaction. 
Our approach strikes a balance between exploration and control: by treating the model's generated images as intermediate outputs and allowing users to select and composite across them, users can take advantage of the model's generative capabilities and use it as an exploration tool, while also retaining fine-grained control over the final result. 
This is especially helpful in cases where users may not know what they want until they see it. 
Second is the ability to correct undesired artifacts in resulting images. 
With our method, 
users can replace undesired regions with more visually appealing regions from other images and build towards their desired result. Compared to the trial-and-error process,  where users ``re-roll the dice'' in hopes of getting a satisfactory image, our approach combinatorially improves the chances of success: users can combine a few images, each one containing a good region.

We show visually compelling results on various applications and user workflows, including creating new appearance combinations, correcting shape misalignment, reducing artifacts, and improving prompt alignment.  Our method outperforms existing blending methods in preserving the fidelity of local regions while maintaining overall realism.
Our code and data are available on our \href{https://lseancs.github.io/generativephotomontage/}{webpage}. We include a breadth of results and ablation experiments in the Appendix.

\begin{figure*}
\centering
  \includegraphics[width=0.8\textwidth]{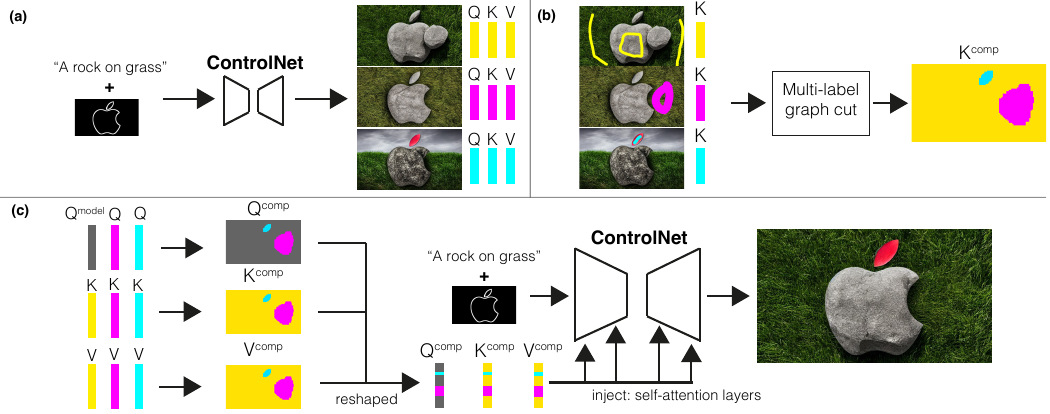}
    \vspace{-10pt}

  \caption{{\bf Overview.} (a) ControlNet-generated images using the same prompt and sketch with different seeds. (b) Upon inspecting the stack, the user wishes to remove the extra rock from the first image and add in the red leaf from the third image. The user draws strokes to select desired regions from each image. Our method finds a segmentation across the stack by performing multi-label graph cut in diffusion feature space ($K$ features). (c) The graph-cut result is then used to form composite $Q, K, V$ features, which are then injected into the self-attention layers. The final result is a harmonious composite of the user-selected regions. }
  \label{fig:method}
  \vspace{-10pt}
\end{figure*}

\section{Related Work}

\myparagraph{Text-to-image generative models} aim to learn the real image distribution conditioned on textual inputs~\cite{zhu2019dm,tao2020df,nichol2021glide}. In recent years, we have seen rapid progress with works on different training objectives, such as diffusion~\cite{sohl2015deep,ho2020denoising,song2020score,karras2022elucidating,karras2023analyzing}, GANs~\cite{sauer2023stylegan,kang2023scaling,sauer2023adversarial}, and autoregressive models~\cite{yu2022scaling,chang2023muse}, as well as new architectures~\cite{peebles2023scalable,dhariwal2021diffusion,Rombach_2022_CVPR,esser2024scaling}. %
However, these models may still fall short of generating what the user wants in one go and often fail to follow all instructions in the text prompt, despite recent efforts~\cite{chefer2023attend,feng2022training,liu2022compositional}. In this work, we aim to bridge this gap by allowing users to compose desirable regions from multiple generated images.

\myparagraph{Image editing.} Text-to-image diffusion models have enabled various editing tasks given reference images and text-instructions~\cite{meng2021sdedit,hertz2024style,huang2024creativesynth,gu2024swapanything,yang2023paint,ye2023ip,zhao2024uni}. This is usually achieved through fine-tuning the model~\cite{kawar2023imagic,brooks2023instructpix2pix} or modifying the denoising process of the diffusion model~\cite{meng2021sdedit,hertz2022prompt,parmar2023zero,tumanyan2022plugandplay,cao2023masactrl,alaluf2023cross, pfbdiff}. 
These methods often focus on the attention mechanism in the text-to-image model, which is crucial for determining the structure and text alignment of generated images~\cite{hertz2022prompt, patashnik2023localizing}.  
While we take inspiration from these works, our tasks and methods are different. Notably, MasaCtrl~\shortcite{cao2023masactrl}, Cross-Image Attention~\shortcite{alaluf2023cross}, and StyleAligned~\shortcite{hertz2024style} focus on high-level style transfer, where local appearances are expected to change. We focus on blending a multi-image stack and preserving local appearances for greater user control.
We show that our proposed technique performs better for this new task. 

\myparagraph{Controllable image generation.} Improving controllability and adherence to text instructions is critical for using these models as a collaborative tool. As a result, many recent works increase user control in the form of input conditions, such as sketch, depth map, bounding box, segmentation map, and reference image~\cite{zhang2023adding, phung2023grounded, avrahami2023spatext, li2023gligen, kim2023dense,Zheng_2023_CVPR,parmar2024one, ma2024subject, gu2024filter, bhat2024loosecontrol}. Another line of work improves the existing text conditioning~\cite{chefer2023attend,feng2022training, bao2024separate}, constrains internal features \cite{tewel2024training, avrahami2024chosen}, or augments it through a rich text editor~\cite{ge2023expressive}. However, these works aim to create a single correct image by directly constraining the output space of solutions. In contrast, we offer a complementary approach -- we allow users to pick and choose exactly what they want from multiple generated images, giving them more fine-grained control without relying solely on the model to create a single perfect image in one shot.

\myparagraph{Image blending} aims to combine multiple images in a seamless manner~\cite{burt1987laplacian,perez2003poisson,farbman2009coordinates,szeliski2011fast}.  Our work draws inspiration from Interactive Digital Photomontage~\cite{agarwala2004interactive}, a seminal work that employs graph cut~\cite{boykov2001fast, boykov2004experimental, kolmogorov2004energy,rother2004grabcut} for blending multiple images given sparse user strokes. This method allows us to ``capture the moment''~\cite{cohen2006moment} or create new visual effects. Many other works also use graph-cut optimization for textures~\cite{kwatra2003graphcut} and videos~\cite{rubinstein2008improved}. Our method follows these graph-cut frameworks but performs the optimization in diffusion feature space, which captures more semantic information compared to pixel colors or edges. 
More recent image blending methods use generative models like GANs~\cite{wu2019gp,zhang2020deep} or diffusion models~\cite{avrahami2023blended, avrahami2023blended, bar2023multidiffusion,sarukkai2024collage,shirakawa2024noisecollage,lee2023syncdiffusion, objectstitch}. However, our method is specifically designed for compositing a spatially aligned image stack and is better at preserving user-selected regions and 
 blending them harmoniously. We also provide additional support for multi-image segmentation with sparse user strokes.

\section{Method} 
Our method takes in a stack of generated images and produces a final image based on sparse user strokes. 
In our image stack, images are generated through ControlNet \cite{zhang2023adding}, using one or more prompts (Figure \ref{fig:method}a). The generated images share common spatial structures, as they are produced using the same input condition (e.g., edge maps or depth maps).

Upon browsing the image stack, the user selects desired objects and regions via broad brush strokes on the images. For example, in Figure~\ref{fig:method}b, the user wishes to remove the rock at the Apple bite in the first image and add the red leaf from the third image. To do so, the user draws strokes on the base rock in the first image, the patch of grass in the second image, and the red leaf in the third image. Our algorithm takes the user input and performs a multi-label graph cut optimization in feature space to find a segmentation of image regions across the stack that minimizes seams. Finally, using a new injection scheme, our method composites the segmented regions during the denoising process (Figure~\ref{fig:method}c).
The final composite image seamlessly blends the user-selected regions while faithfully preserving the local appearances.

 Below, we first give a brief overview of image space graph-cut segmentation in Section~\ref{sec:graphcut}, and then introduce our feature-based multi-image segmentation (Section~\ref{sec:segmentation}) and blending algorithms (Section~\ref{sec:composition}) in more detail.

\subsection{Preliminaries: Segmentation with Graph Cut}
\label{sec:graphcut}

Graph cut~\cite{boykov2001fast, boykov2004experimental, kolmogorov2004energy} has been widely used in several image synthesis and analysis tasks, including texture synthesis~\cite{kwatra2003graphcut}, image synthesis~\cite{agarwala2004interactive}, segmentation~\cite{rother2004grabcut}, and stereo~\cite{vogiatzis2005multi}.  

Here, we describe multi-label graph cut in image space~\cite{agarwala2004interactive,boykov2001fast}. Suppose we have an image stack of $N$ images, labeled $1$ to $N$. For each 2D pixel location $p$ in the output image $I_o$, the goal is to assign an image label $i \in [1...N]$. If a pixel $I_o(p)$ in the output image is assigned the image label $i$, then ${I_o(p) = I_i(p)}$.
The optimization seeks to find an optimal image label assignment for all output pixels such that a given energy cost function is minimized. We can define the energy cost function to encourage the label assignments to have desired properties, such as placing seams in less noticeable regions. An output image of size $(W, H)$ means the optimization has to solve for $W\times H$ variables, where each variable has $N$ candidate labels. To solve the optimization, researchers have adopted max-flow min-cut algorithms for binary cases ($N=2$)~\cite{ford1957simple},  or $\alpha$-expansion for multi-label cases ($N >2$) \cite{boykov2001fast}.

\subsection{Segmentation with Feature-Space Graph Cut}
\label{sec:segmentation}

Given user strokes, our goal is to find a segmentation across the generated image stack and select image regions that adhere to the user strokes while minimizing seams. To achieve this, we also employ a multi-label graph cut optimization. 

\begin{figure}
  \includegraphics[width=\linewidth]{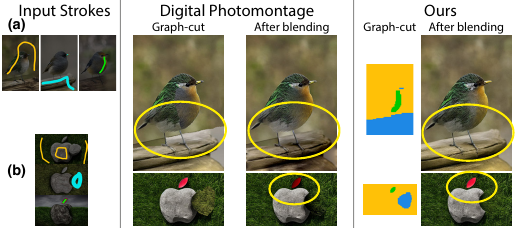}
  \caption{{\bf Ours vs. Interactive Digital Photomontage \cite{agarwala2004interactive}}. 
  (a) Pixel-space graph cut may be more sensitive to low-level changes in color, whereas our diffusion feature-based graph cut selects seams that are more aligned with semantic features. (b) Due to large variations in color across the stack, gradient-domain blending~\cite{perez2003poisson} may alter local appearances, such as the red leaf.}
  \label{fig:dp_ours}
  \vspace{-10pt}
\end{figure}

However, in contrast to prior image-space graph cut approaches~\cite{agarwala2004interactive,boykov2001fast}, we perform the optimization in feature space, using the key features $K \in \mathbb{R}^{w\times h \times d}$ from the self-attention layers of the diffusion model. 
$K$ serves as a compact, lower-resolution representation of the generated image, where $(w, h)$ are smaller than the original image resolution $(W, H)$ and $d$ is the number of hidden dimensions. Prior works show that these features capture rich appearance and semantic information of the generated image~\cite{alaluf2023cross, cao2023masactrl}, which makes them better candidates than raw pixel values for finding good seams. Moreover, subsequent blending in feature space gives us more natural and seamless composites than in pixel space, without additional post-processing as in Agarwala et al.~\shortcite{agarwala2004interactive}. 
See Figure \ref{fig:dp_ours} for comparison.

Instead of solving for each pixel location $(W, H)$, our optimization assigns a label $i\in [1...N]$ to each spatial location $p = (x,y)$ in key features $K$, for $x\in\{1...w\}$ and $y\in\{1...h\}$. During the blending stage, we use the label assignment to create composite self-attention features, which are then injected into ControlNet~\cite{zhang2023adding} to form the final result (Section \ref{sec:composition}).

As users generate the initial image stack, we store the query, key, and value features $Q, K, V$ of each image for all layers and time steps on disk.  
After the user marks desired image regions with strokes, our system performs a multi-label graph cut using the stored key features $K$ from the first encoding layer, where $(w, h) = \frac{1}{8} (W, H)$,   at the final time step, when the generated content is mostly formed.

\myparagraph{Energy cost.} To create a good composite, we design the energy cost function to ensure that the label assignment 1) satisfies user-designated strokes and 2) picks good (unnoticeable) seams to join regions from different images. The energy function is composed of unary and pairwise costs~\cite{boykov2001fast, boykov2004experimental, kolmogorov2004energy}:

\begin{equation}
    E_{\text{total}}(L) =  \sum_{p} E(p, L_p) + \sum_{p, q} E(p, q, L_p, L_q),
\end{equation}
where $p$ and $q$ are neighboring spatial locations in key features $K$. $L_p$ and $L_q$ are image labels to be optimized. 

\myparagraph{Unary term.} Unary costs are the cost of assigning a label $L_p$ at feature location $p$. Here, we assign a high penalty if there is a user stroke at the corresponding pixel location of $p$, and the label is not the image that the user has designated:
\vspace{-5pt}
\begin{equation}
    E(p, L_p) = 
    \begin{cases}
      C & \text{if $S(p, i) = 1$ and $L_p \neq i$} \\
      0 & \text{otherwise},
    \end{cases}
\end{equation}
where $S(p, i)$ is an indicator function of whether there is a user stroke at the corresponding pixel location of $p$ in image $i$, and $C$ is a large constant. We use $C=10^6$ in our results.

\myparagraph{Pairwise term.} Pairwise costs are the cost of assigning neighboring feature locations a pair of labels. Because we want seams to be less noticeable, we encourage seams to fall on edges (lower cost), where the neighboring features are significantly different:
\vspace{-8pt}
\begin{equation}
    E(p, q, L_p, L_q) = 
    \begin{cases}
        \sum_{i=1}^{N} \lambda e^{-\frac{|f_i(p) - f_i(q)|}{2\sigma}} & \text{if $L_p \neq L_q$} \\
        0 & \text{otherwise},
    \end{cases}
    \label{eq:pairwise}
\end{equation}
where $f_i(p)$ is a feature vector derived from the key features $K$ of image $i$ at location $p$. To capture the most important features, $f_i(p)$ consists of the top-$10$ PCA components of $K$ at location $p$, computed across hidden dimensions and heads. 

The cost is low if the features of $p$ and $q$ are dissimilar for all images within the stack; in other words, we encourage seams where $p$ and $q$ straddle an edge in all the images. $\sigma$ controls how quickly the penalty falls off as feature distance increases, and $\lambda$ is a constant scale for the cost range. We use $\lambda=100$ and $\sigma=10$ in all our results.

\myparagraph{Discussion.} While it is possible to segment each image individually using off-the-shelf methods (e.g., SAM \cite{kirillov2023segment}), it often requires extra user intervention to resolve conflicts where objects overlap and to maintain coverage. Instead, our multi-label graph cut optimization can automatically account for all images within the stack, assigning a unique label to each location. %

\subsection{Composition with Self-Attention Feature \\ Injection} \label{sec:composition}

The above optimization gives us an image label assignment per feature location for the output image. We use this assignment to make composite features $Q^{\text{comp}}$, $K^{\text{comp}}$ and $V^{\text{comp}}$ from the respective features $Q, K, V$ of the image stack, for each self-attention layer in ControlNet. We then inject these composite features $Q^{\text{comp}}$, $K^{\text{comp}}$ and $V^{\text{comp}}$ into ControlNet to form the final result.

When users input strokes, they designate one image within the stack as the base image (usually the one whose background region is selected). During the blending stage, we use the seed and prompt of this base image when injecting the composite features.

Specifically, we resize the label assignment map $L$ into the respective sizes of each self-attention layer $l$. Then, we make composite features $Q^{\text{comp}}$, $K^{\text{comp}}$ and $V^{\text{comp}}$ as follows:
\vspace{-15pt}
\begin{align}
    Q^{\text{comp}}_{l} &= M^B_l \odot Q_l^{\text{model}} + \sum_{i \neq B} M^i_l \odot Q^{i}_{l}, \; \label{eq:q_comp} \\
    K^{\text{comp}}_{l} &= \sum_i M^i_l \odot K^{i}_{l}, \; \label{eq:k_comp} \\
    V^{\text{comp}}_{l} &= \sum_i  M^i_l \odot V^{i}_{l}, \; \label{eq:v_comp}
\end{align}
where $M^i_l$ is the binary mask of feature locations with the label assignment $i$, resized to layer $l$. $M^B$ represents the mask of the base image, where $B$ is the base image index. $Q^{i}_{l}$ are the query features $Q$ from image $i$ at layer $l$ (stored during initial generation), and similarly for $K^{i}_{l}$ and $V^{i}_{l}$. $Q^{\text{model}}$ are the query features generated from the model during the blending stage, which are different from those during the initial generation. These composite  features are injected into the U-Net's self-attention maps for all layers and time steps.

\begin{figure}
  \includegraphics[width=\linewidth]{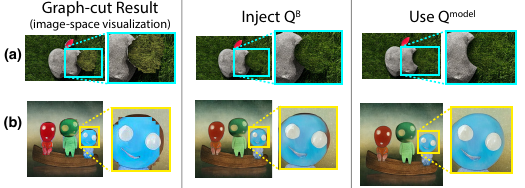}
   \vspace{-15pt}
  \caption{\textbf{Using $Q^B$ vs. $Q^{\text{model}}$}. $Q^B$: query features of the base image from initial generation. $Q^{\text{model}}$: query features generated from the model during the blending stage. Leftmost column: visualization of diffusion-feature graph cut results, resized to image dimensions and combined in image space. (a) Injecting $Q^B$ does not leave room for the model to adapt its image structure near seams, causing the shadow from the input image to remain. (b) Injecting $Q^B$ causes the image to strictly adhere to low-resolution graph cut boundaries, whereas using $Q^{\text{model}}$ adjusts the boundaries to align with semantic features in high-resolution image space.}
  \label{fig:fixedQ}
   \vspace{-10pt}
\end{figure}

Note that we inject the initially generated self-attention features for all images \emph{except} for $Q^B$, the query features of the base image. %
If we inject the initial $Q^B$ features, we often observe suboptimal blending at the seams.
As noted in previous literature \cite{alaluf2023cross, cao2023masactrl}, $Q$ influences the image structure, while $K$ and $V$ influence the appearance. Hence, injecting $Q^B$ (and thus completely overwriting the $Q$ features) eliminates the opportunity for the model to adapt the image structure near the seams. 
Allowing $Q$ within the mask $M^B$ to change over time allows the model to adapt to the different graph-cut regions when blending (Figure \ref{fig:fixedQ}a). It also adjusts the low-resolution graph cut boundaries to align with semantic features in high-resolution pixel space (Figure \ref{fig:fixedQ}b).

\vspace{5pt}
\section{Results} \label{sec:results}

\begin{figure*}[h]
  \includegraphics[width=\textwidth]{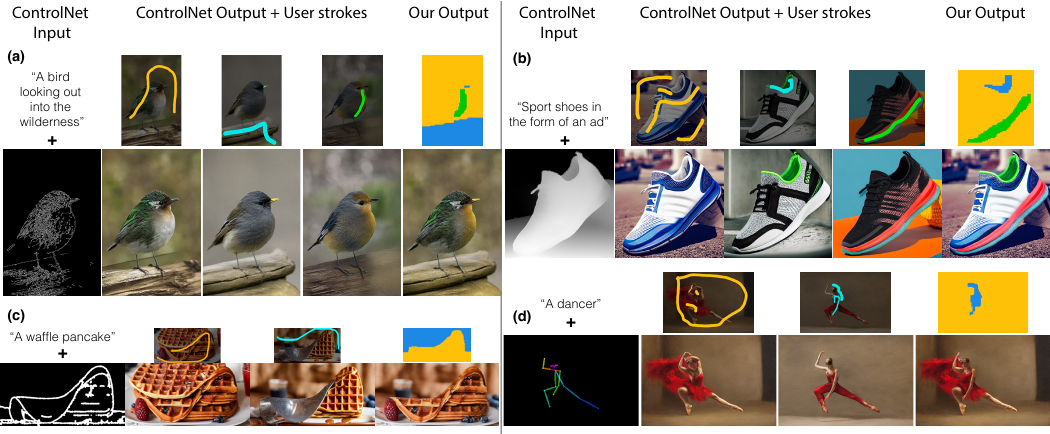}
  \caption{\textbf{Appearance Mixing and Shape Correction}. Our method can be used to mix appearances for creative exploration and design (a, b). Our method can also fix incorrect shapes and artifacts from ControlNet's outputs (c, d) , which often occur for uncommon input shapes.
  }
  \label{fig:results_appearance_shape}
  \vspace{-5pt}
\end{figure*}

\begin{figure*}[h]
    \centering
  \includegraphics[width=0.9\linewidth]{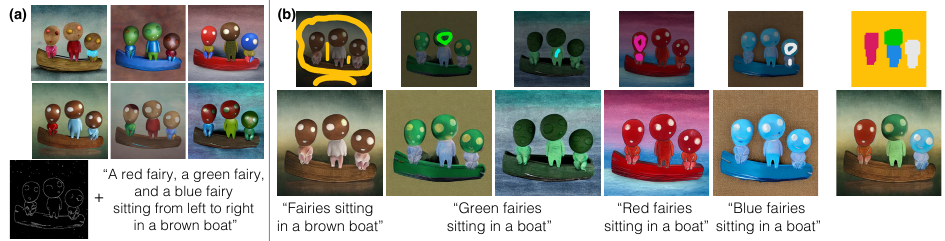}
  \caption{\textbf{Prompt Alignment}. Our method can be used to increase alignment to long, complicated prompts. (a) Example where vanilla ControlNet's outputs do not adhere to the prompt. (b) With Generative Photomontage, users can create the desired image by combining the outputs from shorter, simpler prompts.}
  \label{fig:fairies}
  \vspace{-10pt}
\end{figure*}

By supporting the ability to combine generated images, our method allows users to achieve a wider range of results with more flexibility and control. %
Here, we highlight some use cases and show compelling results for each application.
Due to space constraints, we refer readers to the Appendix for additional results. We created the results using various pre-trained ControlNet models~\cite{zhang2023adding} with Stable Diffusion 1.5, such as canny edge, scribble map, Openpose~\cite{openpose}, and depth map.

\myparagraph{Appearance Mixing.} First, we show applications in creative and artistic design, where users refine images based on subjective preference. This is useful in cases where the user may not realize what they want until they see it (e.g., creative exploration).
For example, users may use our method for exploring architectural designs, by combining the roofs, windows, and doors from different images (Figure \ref{fig:teaser}c), 
or for creating fashion designs, by mixing different features of shoes (Figure \ref{fig:results_appearance_shape}b).
Additionally, users can composite different components to create something new. For example, we can combine the body, ear, and arm of a robot to form a new robot (Figure \ref{fig:teaser}a). This strategy can also be applied to other subjects, e.g., to combine new colors in a bird's feathers (Figure \ref{fig:results_appearance_shape}a).

\myparagraph{Shape and Artifacts Correction.} While users can provide a sketch to guide ControlNet's output, ControlNet may fail to adhere to the user's input condition, especially when asked to generate objects with uncommon shapes. In such cases, our method can be used to ``correct'' object shapes and scene layouts, given a replaceable image patch within the stack. 

For example, suppose the user wishes to create an Apple-logo-shaped rock and prompts ControlNet with ``A rock on grass'' alongside an Apple-logo sketch. Since Apple-logo-shaped rocks are not commonly seen in real life (and thus out-of-training distribution), the model fails to produce the desired image. Figure \ref{fig:method} shows an additional rock piece covering the apple bite. To correct it, the user can use our framework to replace it with a patch of grass from another image in the stack (Figure \ref{fig:method}b). Similarly, users can correct ControlNet's output to create waffles in the shape of famous architectural buildings (Figure \ref{fig:results_appearance_shape}c), which is difficult to achieve with ControlNet alone due to the rare object-shape combination. 
Finally, we show other correction examples, such as correcting a dancer's pose (Figure \ref{fig:results_appearance_shape}d),
or replacing an unrealistic-looking dog with a different one (Figure \ref{fig:teaser}d).

\myparagraph{Prompt Alignment.} In addition, our method can be used to increase prompt alignment in cases where the generated output does not accurately follow the input prompt. For example, text-to-image with ControlNet can often fail to follow all aspects of a long complicated prompt, such as ``A red fairy, a green fairy, and a blue fairy sitting from left to right in a brown boat'' (Figure \ref{fig:fairies}a). Using our method, users can create the desired image by breaking it up into simpler prompts and selectively combining the outputs (Figure \ref{fig:fairies}b).  
Since our results depend on the availability of at least one ``correct'' candidate per region within the generated stack, we encourage our method to be used in conjunction with existing methods \cite{chefer2023attend, feng2022training} for greater accuracy and control.

\section{Evaluation}
\label{sec:evaluation}

For a comprehensive evaluation, we created $20$ test examples, spanning the use cases in Section \ref{sec:results}. Below, we compare our method against several baselines. We include the full list of test examples, additional details, and ablations in the Appendix.

\begin{figure*}
\centering
  \includegraphics[width=0.95\textwidth]{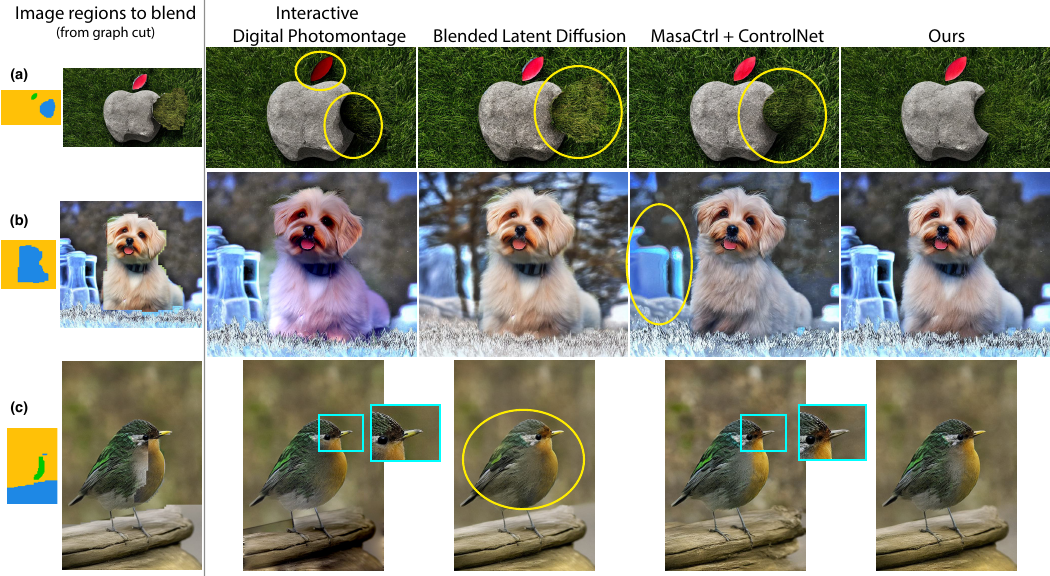}
  \caption{\textbf{Qualitative Comparison}. Leftmost column: Image regions to blend (output of graph-cut optimization). Graph-cut in diffusion feature space is visualized on the left, and the image-space 
  composite of that graph-cut is visualized on the right.
  Interactive Digital Photomontage \cite{agarwala2004interactive}: its pixel-space graph-cut may cause seams to fall on undesired edges (see Figure \ref{fig:dp_ours} also), and their gradient-domain blending often fails to preserve color. Blended Latent Diffusion \cite{avrahami2023blended} and MasaCtrl+ControlNet \cite{cao2023masactrl} may lead to color and structure changes.}
  \label{fig:qualitative}
  \vspace{-10pt}
\end{figure*}

\myparagraph{Baselines.} We select baselines that perform blending in pixel space:  Interactive Digital Photomontage (IDP) \cite{agarwala2004interactive}; noise space: Blended Latent Diffusion (BLD) \cite{avrahami2023blended}; and attention space: MasaCtrl with ControlNet \cite{cao2023masactrl}. We also compare with CollageDiffusion \cite{sarukkai2024collage}, Cross-Domain Compositing (CDC) \cite{hachnochi2023crossdomain}, GP-GAN \cite{wu2019gp}, and Deep Image Blending \cite{zhang2020deep}. In addition, we compare with a modified version of BLD with greater noise overlap, inspired by MultiDiffusion \cite{bar2023multidiffusion}. Please see the Appendix for implementation details. When running comparisons, we use user strokes as input for IDP and our graph cut masks as input for the other baselines.

\begin{table}[!t]
\renewcommand{\arraystretch}{0.85}
\setlength{\tabcolsep}{5pt}
\resizebox{\linewidth}{!}{
\begin{tabular}{ l c c c c} 
\toprule
& \shortstack[c]{\textbf{Masked} \\ \textbf{LPIPS}}$\downarrow$ & 
\shortstack[c]{\textbf{Masked} \\ \textbf{SSIM}}$\uparrow$ &
\textbf{PSNR} $\uparrow$ & 
\shortstack[c]{\textbf{Seam Gradient Score} \\
\footnotesize{min, avg, max} \\
\footnotesize{0.255, 0.340, 0.425} 
} \\
\midrule
\textbf{Ours} & \underline{0.123} &  0.815 & \textbf{22.46} & 0.339 \\
IDP \cite{agarwala2004interactive} & \textbf{0.104} & \textbf{0.888} & 20.13  & 0.306 \\
BLD \cite{avrahami2023blended} & 0.222 & 0.772 & 20.27 & 0.393  \\
BLD+Multi \cite{bar2023multidiffusion} & 0.224 &  0.766 & 19.79 & 0.318 \\
MasaCtrl+CtrlNet \cite{cao2023masactrl} & 0.230 & 0.680 & 18.34 & 0.341 \\
Deep Img Blending \cite{zhang2020deep} & 0.270 & 0.766 & 17.33 & 0.313 \\
GP-GAN \cite{wu2019gp} & 0.226 & \underline{0.820} & 17.45 & 0.220* \\
CDC \cite{hachnochi2023crossdomain} & 0.376 & 0.584 & 20.01 & 0.456* \\
CollageDiffusion \cite{sarukkai2024collage} & 0.243 & 0.605 & \underline{20.57} & 0.559* \\
\bottomrule
\end{tabular}
}
\caption{\textbf{Quantitative Results}. 
To measure fidelity of local image regions, we report masked LPIPS and SSIM and PSNR. 
To measure blending quality of seams, we compute a seam gradient (SG) score, which is the mean gradient magnitude along seams that join different image regions. 
For reference, we also compute the minimum, average, and maximum SG scores of each image stack and report their averages on the top.
Methods with SG scores that fall outside of this range (*) exhibit significant seam artifacts. 
}
\label{table:quantitative}
\vspace{-15pt}
\end{table}

\myparagraph{Quantitative Metrics.} 
To measure local appearance fidelity, we computed masked LPIPS \cite{zhang2018perceptual}, masked SSIM \cite{wang2004image}, and PSNR \cite{hore2010image}.
For LPIPS and SSIM, we use the masks from our feature-based graph cut, resized to image dimensions. 
For PSNR, we compare each blended result with the image-space composite (also created with the graph cut masks). Note that the image-space composites contain noticeable seams (e.g., the first column in Figure \ref{fig:qualitative}) and only serve as a proxy in the absence of groundtruth data. 

To measure seam artifacts, we compute the average gradient magnitude along seams, called the seam gradient (SG) score. Since some seams fall on object boundaries, the gradient is not expected to be zero. Rather, it should be within range of the SG scores of images in the stack. For reference, we compute the minimum, average, and maximum SG scores of each image stack and report their averages in Table \ref{table:quantitative}.

\myparagraph{Quantitative Results.}  As shown in Table \ref{table:quantitative}, our method achieves the highest PSNR score, the second-best LPIPS loss, and the third-best SSIM score across all the baselines, indicating good preservation of local image appearance. While IDP achieved the best LPIPS and SSIM scores, it often exhibits significant color changes (e.g., Figure \ref{fig:qualitative}), leading to a lower PSNR score than ours. Our method's SG score is within range and close to the average SG score.
Though CollageDiffusion and GP-GAN achieved the second-best PSNR and SSIM, respectively, they have out-of-range SG scores, denoting major seam artifacts.
Deep Image Blending and Cross-Domain Compositing have relatively large LPIPS losses, which correlates with larger changes in image appearances.
BLD has the third-highest PSNR score and outperforms its variant, BLD+MultiDiffusion. Please see the Appendix for examples.

\begin{figure}
  \centering
  \includegraphics[width=\linewidth]{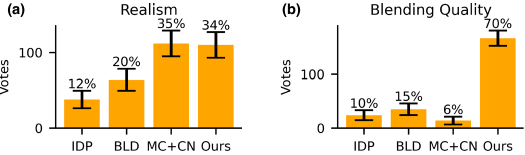}
  \vspace{-20pt}
  \caption{{\bf User Survey Results}. For each test scene, we asked participants to select the best image produced by four methods. The results show that our method has the best blending quality while being comparable to MasaCtrl+CtrlNet in realism. Error bars: SE.
  }\label{fig:survey}
  \vspace{-5pt}
\end{figure}

\myparagraph{Qualitative Comparison.} In Figure \ref{fig:qualitative}, we show qualitative comparisons with
IDP, BLD, and MasaCtrl+ControlNet. As shown, IDP often leads to changes in color due to gradient-domain blending, such as the marked regions in Figure \ref{fig:qualitative}. 
BLD \cite{avrahami2023blended} and MasaCtrl+ControlNet \cite{cao2023masactrl} may lead to structural changes or local appearance changes. The other baselines exhibit significant artifacts in blended regions or at seams.
Please see the Appendix for qualitative results of these other baselines.

\myparagraph{User Survey.} Finally, we conduct two user surveys to compare our results with the most competitive baseline in each domain: Interactive Digital Photomontage (pixel-space), BLD (noise-based), and MasaCtrl+ControlNet (attention-based). Across $12$ test scenes, we ask participants to select the best image produced by the four methods: (1) Which image appears most realistic to you? and (2) Which image is the best at blending all the selected regions? Out of $324$ and $240$ responses, respectively, results show that our method is comparable to MasaCtrl+ControlNet in terms of realism and has the best blending quality among the baselines by a wide margin (Figure \ref{fig:survey}).

\subsection{Ablation}
\label{sec:ablation_main}

\begin{table}[!t]
\centering
\renewcommand{\arraystretch}{0.8}
\setlength{\tabcolsep}{5pt}
\resizebox{\linewidth}{!}{
\begin{tabular}{ l c c c c} 
\toprule
& \shortstack[c]{\textbf{Masked} \\ \textbf{LPIPS}}$\downarrow$ & 
\shortstack[c]{\textbf{Masked} \\ \textbf{SSIM}}$\uparrow$ &
\textbf{PSNR} $\uparrow$ & 
\shortstack[c]{\textbf{Seam Gradient Score} \\
\footnotesize{min, avg, max} \\
\footnotesize{0.255, 0.340, 0.425} 
} \\
\midrule
 \textbf{Ours} & \textbf{0.123} &  \textbf{0.815} & \textbf{22.46} & 0.339\\
 w/ $K^{\text{concat}}$, $V^{\text{concat}}$ &  0.243 & 0.677 & 18.37 & 0.354\\
 w/ $K^{\text{model}}$, $V^{\text{model}}$ & 0.268 & 0.669 & 18.85 & 0.332\\
\bottomrule 
\end{tabular}
}
\vspace{-10pt}
\caption{{\bf Ablation of Self-Attention Injection Schemes.}
}
\label{table:ablation_injection}
\vspace{-10pt}
\end{table}

Here, we ablate our self-attention injection scheme with alternative injection strategies, adapted to our use case. 
First, we consider using shared (concatenated) key and value features, i.e, $K^{\text{concat}} =[K^1, K^2, ...,K^N]$ and $V^{\text{concat}} = [V^1, V^2, ...,V^N]$, which StyleAligned~\shortcite{hertz2024style} used to transfer style across different images. 
Next, we consider using $K^{\text{model}}$ and $V^{\text{model}}$ for the base image, similar to Equation \ref{eq:q_comp}. 
Both alternative injection schemes lead to changes in the appearance of user-selected image regions, whereas our method is able to preserve them. Table \ref{table:ablation_injection} reflects this change quantitatively.
Please see the Appendix for qualitative examples and additional ablations on attention injection.

\subsection{Multi-image Segmentation}
\label{sec:multi_img_seg}

\begin{figure}
\centering
  \includegraphics[width=0.85\linewidth]{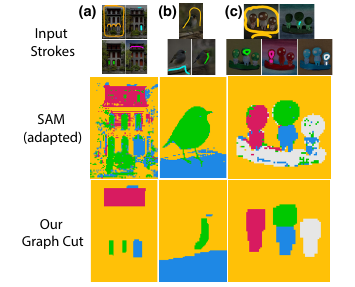}
  \vspace{-10pt}
  \caption{\textbf{Our Graph Cut vs. SAM \cite{kirillov2023segment}}. 
  We adapted SAM to an image stack and compared it with our graph cut.
  As shown, SAM may output noisy, incongruent labels within an object, such as the building (a) and fairies (c). SAM may also fail to follow user strokes, such as the bird interior (b) and the boat (c). Our graph cut takes into account the entire image stack during optimization and outputs labels that are congruent and satisfy user strokes.}
  \label{fig:sam}
  \vspace{-15pt}
\end{figure}

Here, we compare our multi-label graph cut with a modified version of SAM \cite{kirillov2023segment} and pixel-space graph cut in IDP \cite{agarwala2004interactive}. 
Since SAM is trained to segment single images, adapting it to a multi-image stack while maintaining coverage and avoiding overlaps is not straightforward. As a baseline, we use the following setup: given an input image stack, we run SAM on each image, with the user strokes on the image as positive and strokes on other images as negative samples. SAM outputs logits per pixel per image. For each pixel in the composite image, we assign the image label with the highest logit score. As shown in Figure \ref{fig:sam}, we observe two major types of artifacts here: 1) incongruent (noisy) labels within an object; 2) the segmentation does not follow user strokes.
Pixel-space graph cut, as shown in Figure~\ref{fig:dp_ours},  may be
sensitive to low-level changes in color and select seams that fall on undesired edges. Our graph cut, which operates in feature space, selects seams that are more aligned with semantic features. 
Please see the Appendix for ablations of these methods and other graph cut features.

\section{Discussion}

In this work, 
we proposed a new approach for generating images: 
by compositing it from multiple ControlNet-generated images. 
At a broader level, our work suggests a new user workflow for interacting with text-to-image models: rather than trying to get the model to output the final end-product (i.e., a \textit{single} image that contains everything the user wants, which is often difficult), we treat the model's output as an \textit{intermediate} step, from which users provide further input to create their final end-product. %
Our approach not only gives users more fine-grained control over the final output, but also allows us to fully utilize the model's generative capabilities in creating diverse candidates. 
We hope this work inspires new ways of interacting with generative models.

\myparagraph{Acknowledgments}
We are grateful to Kangle Deng for his help with setting up the user survey. We also thank Maxwell Jones, Gaurav Parmar, Sheng-Yu Wang, and Or Patashnik for helpful comments and suggestions. This project is partly supported by the Amazon Faculty Research Award, DARPA ECOLE, the Packard Fellowship, the IITP grant funded by the Korean Government (MSIT) (No. RS-2024-00457882, National AI Research Lab Project), and a joint NSFC-ISF Research Grant no. 3077/23.

{
    \small
    \bibliographystyle{ieeenat_fullname}
    \bibliography{citations}

\begin{thebibliography}{95}
\providecommand{\natexlab}[1]{#1}
\providecommand{\url}[1]{\texttt{#1}}
\expandafter\ifx\csname urlstyle\endcsname\relax
  \providecommand{\doi}[1]{doi: #1}\else
  \providecommand{\doi}{doi: \begingroup \urlstyle{rm}\Url}\fi

\bibitem[Agarwala et~al.(2004)Agarwala, Dontcheva, Agrawala, Drucker, Colburn, Curless, Salesin, and Cohen]{agarwala2004interactive}
Aseem Agarwala, Mira Dontcheva, Maneesh Agrawala, Steven Drucker, Alex Colburn, Brian Curless, David Salesin, and Michael Cohen.
\newblock Interactive digital photomontage.
\newblock In \emph{ACM SIGGRAPH}, 2004.

\bibitem[Alaluf et~al.(2024)Alaluf, Garibi, Patashnik, Averbuch-Elor, and Cohen-Or]{alaluf2023cross}
Yuval Alaluf, Daniel Garibi, Or Patashnik, Hadar Averbuch-Elor, and Daniel Cohen-Or.
\newblock Cross-image attention for zero-shot appearance transfer.
\newblock In \emph{ACM SIGGRAPH}, 2024.

\bibitem[Auguste08(2023)]{fairy_source}
Auguste08.
\newblock Kodama princess mononoke 3d print model.
\newblock \href{https://www.cgtrader.com/3d-print-models/art/sculptures/kodama-princess-mononoke-88ce8f61-6446-4178-9882-8cae128868ac}{See source}, 2023.

\bibitem[Avrahami et~al.(2023{\natexlab{a}})Avrahami, Fried, and Lischinski]{avrahami2023blended}
Omri Avrahami, Ohad Fried, and Dani Lischinski.
\newblock Blended latent diffusion.
\newblock \emph{ACM Transactions on Graphics (TOG)}, 42\penalty0 (4), 2023{\natexlab{a}}.

\bibitem[Avrahami et~al.(2023{\natexlab{b}})Avrahami, Hayes, Gafni, Gupta, Taigman, Parikh, Lischinski, Fried, and Yin]{avrahami2023spatext}
Omri Avrahami, Thomas Hayes, Oran Gafni, Sonal Gupta, Yaniv Taigman, Devi Parikh, Dani Lischinski, Ohad Fried, and Xi Yin.
\newblock Spatext: Spatio-textual representation for controllable image generation.
\newblock In \emph{IEEE Conference on Computer Vision and Pattern Recognition (CVPR)}, 2023{\natexlab{b}}.

\bibitem[Avrahami et~al.(2024)Avrahami, Hertz, Vinker, Arar, Fruchter, Fried, Cohen-Or, and Lischinski]{avrahami2024chosen}
Omri Avrahami, Amir Hertz, Yael Vinker, Moab Arar, Shlomi Fruchter, Ohad Fried, Daniel Cohen-Or, and Dani Lischinski.
\newblock The chosen one: Consistent characters in text-to-image diffusion models.
\newblock In \emph{ACM SIGGRAPH}, 2024.

\bibitem[Bao et~al.(2024)Bao, Li, Singh, Wang, and Hebert]{bao2024separate}
Zhipeng Bao, Yijun Li, Krishna~Kumar Singh, Yu-Xiong Wang, and Martial Hebert.
\newblock Separate-and-enhance: Compositional finetuning for text-to-image diffusion models.
\newblock In \emph{ACM SIGGRAPH}, 2024.

\bibitem[Bar-Tal et~al.(2023)Bar-Tal, Yariv, Lipman, and Dekel]{bar2023multidiffusion}
Omer Bar-Tal, Lior Yariv, Yaron Lipman, and Tali Dekel.
\newblock Multidiffusion: Fusing diffusion paths for controlled image generation.
\newblock In \emph{International Conference on Machine Learning (ICML)}, 2023.

\bibitem[Bhat et~al.(2024)Bhat, Mitra, and Wonka]{bhat2024loosecontrol}
Shariq~Farooq Bhat, Niloy Mitra, and Peter Wonka.
\newblock Loosecontrol: Lifting controlnet for generalized depth conditioning.
\newblock In \emph{ACM SIGGRAPH 2024 Conference Papers}, pages 1--11, 2024.

\bibitem[BillionPhotos.com()]{lipstick_source}
BillionPhotos.com.
\newblock A red glitter polish lipstick.
\newblock \href{https://stock.adobe.com/images/a-red-glitter-polish-lipstick/573739903}{See source}.

\bibitem[Boykov and Kolmogorov(2004)]{boykov2004experimental}
Yuri Boykov and Vladimir Kolmogorov.
\newblock An experimental comparison of min-cut/max-flow algorithms for energy minimization in vision.
\newblock \emph{IEEE Transactions on Pattern Analysis and Machine Intelligence (TPAMI)}, 26\penalty0 (9), 2004.

\bibitem[Boykov et~al.(2001)Boykov, Veksler, and Zabih]{boykov2001fast}
Yuri Boykov, Olga Veksler, and Ramin Zabih.
\newblock Fast approximate energy minimization via graph cuts.
\newblock \emph{IEEE Transactions on Pattern Analysis and Machine Intelligence (TPAMI)}, 23\penalty0 (11), 2001.

\bibitem[Brooks et~al.(2023)Brooks, Holynski, and Efros]{brooks2023instructpix2pix}
Tim Brooks, Aleksander Holynski, and Alexei~A Efros.
\newblock Instructpix2pix: Learning to follow image editing instructions.
\newblock In \emph{IEEE Conference on Computer Vision and Pattern Recognition (CVPR)}, 2023.

\bibitem[Burt and Adelson(1987)]{burt1987laplacian}
Peter~J Burt and Edward~H Adelson.
\newblock The laplacian pyramid as a compact image code.
\newblock In \emph{Readings in computer vision}. 1987.

\bibitem[Cao et~al.(2023)Cao, Wang, Qi, Shan, Qie, and Zheng]{cao2023masactrl}
Mingdeng Cao, Xintao Wang, Zhongang Qi, Ying Shan, Xiaohu Qie, and Yinqiang Zheng.
\newblock Masactrl: Tuning-free mutual self-attention control for consistent image synthesis and editing.
\newblock In \emph{IEEE International Conference on Computer Vision (ICCV)}, 2023.

\bibitem[{Cao} et~al.(2019){Cao}, {Hidalgo Martinez}, {Simon}, {Wei}, and {Sheikh}]{openpose}
Z. {Cao}, G. {Hidalgo Martinez}, T. {Simon}, S. {Wei}, and Y.~A. {Sheikh}.
\newblock Openpose: Realtime multi-person 2d pose estimation using part affinity fields.
\newblock \emph{IEEE Transactions on Pattern Analysis and Machine Intelligence (TPAMI)}, 2019.

\bibitem[Chang et~al.(2023)Chang, Zhang, Barber, Maschinot, Lezama, Jiang, Yang, Murphy, Freeman, Rubinstein, et~al.]{chang2023muse}
Huiwen Chang, Han Zhang, Jarred Barber, AJ Maschinot, Jose Lezama, Lu Jiang, Ming-Hsuan Yang, Kevin Murphy, William~T Freeman, Michael Rubinstein, et~al.
\newblock Muse: Text-to-image generation via masked generative transformers.
\newblock In \emph{International Conference on Machine Learning (ICML)}, 2023.

\bibitem[Chefer et~al.(2023)Chefer, Alaluf, Vinker, Wolf, and Cohen-Or]{chefer2023attend}
Hila Chefer, Yuval Alaluf, Yael Vinker, Lior Wolf, and Daniel Cohen-Or.
\newblock Attend-and-excite: Attention-based semantic guidance for text-to-image diffusion models.
\newblock \emph{ACM Transactions on Graphics (TOG)}, 42\penalty0 (4), 2023.

\bibitem[Cohen and Szeliski(2006)]{cohen2006moment}
Michael~F Cohen and Richard Szeliski.
\newblock The moment camera.
\newblock \emph{Computer}, 39\penalty0 (8), 2006.

\bibitem[Cross(2021)]{red_source}
Heidi Cross.
\newblock \href{https://www.pinterest.com/pin/cool-wallpaper--135671007535844786/}{See source}, 2021.

\bibitem[Dhariwal and Nichol(2021)]{dhariwal2021diffusion}
Prafulla Dhariwal and Alexander Nichol.
\newblock Diffusion models beat gans on image synthesis.
\newblock In \emph{Conference on Neural Information Processing Systems (NeurIPS)}, 2021.

\bibitem[Esser et~al.(2024)Esser, Kulal, Blattmann, Entezari, M{\"u}ller, Saini, Levi, Lorenz, Sauer, Boesel, et~al.]{esser2024scaling}
Patrick Esser, Sumith Kulal, Andreas Blattmann, Rahim Entezari, Jonas M{\"u}ller, Harry Saini, Yam Levi, Dominik Lorenz, Axel Sauer, Frederic Boesel, et~al.
\newblock Scaling rectified flow transformers for high-resolution image synthesis.
\newblock In \emph{International Conference on Machine Learning (ICML)}, 2024.

\bibitem[Farbman et~al.(2009)Farbman, Hoffer, Lipman, Cohen-Or, and Lischinski]{farbman2009coordinates}
Zeev Farbman, Gil Hoffer, Yaron Lipman, Daniel Cohen-Or, and Dani Lischinski.
\newblock Coordinates for instant image cloning.
\newblock \emph{ACM Transactions on Graphics (TOG)}, 28\penalty0 (3), 2009.

\bibitem[Feng et~al.(2023)Feng, He, Fu, Jampani, Akula, Narayana, Basu, Wang, and Wang]{feng2022training}
Weixi Feng, Xuehai He, Tsu-Jui Fu, Varun Jampani, Arjun Akula, Pradyumna Narayana, Sugato Basu, Xin~Eric Wang, and William~Yang Wang.
\newblock Training-free structured diffusion guidance for compositional text-to-image synthesis.
\newblock In \emph{International Conference on Learning Representations (ICLR)}, 2023.

\bibitem[Ford and Fulkerson(1957)]{ford1957simple}
Lester~Randolph Ford and Delbert~R Fulkerson.
\newblock A simple algorithm for finding maximal network flows and an application to the hitchcock problem.
\newblock \emph{Canadian journal of Mathematics}, 9, 1957.

\bibitem[Ge et~al.(2023)Ge, Park, Zhu, and Huang]{ge2023expressive}
Songwei Ge, Taesung Park, Jun-Yan Zhu, and Jia-Bin Huang.
\newblock Expressive text-to-image generation with rich text.
\newblock In \emph{IEEE International Conference on Computer Vision (ICCV)}, 2023.

\bibitem[Gu et~al.(2024{\natexlab{a}})Gu, Wang, Zhao, Xiong, Liu, Zhang, Zhang, Zhang, Jung, and Wang]{gu2024swapanything}
Jing Gu, Yilin Wang, Nanxuan Zhao, Wei Xiong, Qing Liu, Zhifei Zhang, He Zhang, Jianming Zhang, HyunJoon Jung, and Xin~Eric Wang.
\newblock Swapanything: Enabling arbitrary object swapping in personalized visual editing.
\newblock \emph{arXiv preprint arXiv:2404.05717}, 2024{\natexlab{a}}.

\bibitem[Gu et~al.(2024{\natexlab{b}})Gu, Yang, and Davis]{gu2024filter}
Zeqi Gu, Ethan Yang, and Abe Davis.
\newblock Filter-guided diffusion for controllable image generation.
\newblock In \emph{ACM SIGGRAPH}, 2024{\natexlab{b}}.

\bibitem[Guerreiro et~al.(2023)Guerreiro, Nakazawa, and Stenger]{pctnet}
Julian Jorge~Andrade Guerreiro, Mitsuru Nakazawa, and Björn Stenger.
\newblock Pct-net: Full resolution image harmonization using pixel-wise color transformations.
\newblock In \emph{IEEE Conference on Computer Vision and Pattern Recognition (CVPR)}, 2023.

\bibitem[Hachnochi et~al.(2023)Hachnochi, Zhao, Orzech, Gal, Mahdavi-Amiri, Cohen-Or, and Bermano]{hachnochi2023crossdomain}
Roy Hachnochi, Mingrui Zhao, Nadav Orzech, Rinon Gal, Ali Mahdavi-Amiri, Daniel Cohen-Or, and Amit~Haim Bermano.
\newblock Cross-domain compositing with pretrained diffusion models.
\newblock \emph{arXiv preprint arXiv:2302.10167}, 2023.

\bibitem[Hertz et~al.(2023)Hertz, Mokady, Tenenbaum, Aberman, Pritch, and Cohen-Or]{hertz2022prompt}
Amir Hertz, Ron Mokady, Jay Tenenbaum, Kfir Aberman, Yael Pritch, and Daniel Cohen-Or.
\newblock Prompt-to-prompt image editing with cross attention control.
\newblock In \emph{International Conference on Learning Representations (ICLR)}, 2023.

\bibitem[Hertz et~al.(2024)Hertz, Voynov, Fruchter, and Cohen-Or]{hertz2024style}
Amir Hertz, Andrey Voynov, Shlomi Fruchter, and Daniel Cohen-Or.
\newblock Style aligned image generation via shared attention.
\newblock In \emph{IEEE Conference on Computer Vision and Pattern Recognition (CVPR)}, 2024.

\bibitem[Ho et~al.(2020)Ho, Jain, and Abbeel]{ho2020denoising}
Jonathan Ho, Ajay Jain, and Pieter Abbeel.
\newblock Denoising diffusion probabilistic models.
\newblock In \emph{Conference on Neural Information Processing Systems (NeurIPS)}, 2020.

\bibitem[Hore and Ziou(2010)]{hore2010image}
Alain Hore and Djemel Ziou.
\newblock Image quality metrics: Psnr vs. ssim.
\newblock In \emph{International Conference on Pattern Recognition (ICPR)}, 2010.

\bibitem[Huang et~al.(2024)Huang, Dong, Zhang, Tang, Li, Ma, Li, and Xu]{huang2024creativesynth}
Nisha Huang, Weiming Dong, Yuxin Zhang, Fan Tang, Ronghui Li, Chongyang Ma, Xiu Li, and Changsheng Xu.
\newblock Creativesynth: Creative blending and synthesis of visual arts based on multimodal diffusion.
\newblock \emph{arXiv preprint arXiv:2401.14066}, 2024.

\bibitem[Huang et~al.(2023)Huang, Tu, and Xu]{pfbdiff}
Wenjing Huang, Shikui Tu, and Lei Xu.
\newblock Pfb-diff: Progressive feature blending diffusion for text-driven image editing.
\newblock \emph{arXiv preprint arXiv:2306.16894}, 2023.

\bibitem[Johnson et~al.(2016)Johnson, Alahi, and Fei-Fei]{johnson2016perceptual}
Justin Johnson, Alexandre Alahi, and Li Fei-Fei.
\newblock Perceptual losses for real-time style transfer and super-resolution.
\newblock In \emph{European Conference on Computer Vision (ECCV)}, 2016.

\bibitem[Kang et~al.(2023)Kang, Zhu, Zhang, Park, Shechtman, Paris, and Park]{kang2023scaling}
Minguk Kang, Jun-Yan Zhu, Richard Zhang, Jaesik Park, Eli Shechtman, Sylvain Paris, and Taesung Park.
\newblock Scaling up gans for text-to-image synthesis.
\newblock In \emph{IEEE Conference on Computer Vision and Pattern Recognition (CVPR)}, 2023.

\bibitem[Karras et~al.(2022)Karras, Aittala, Aila, and Laine]{karras2022elucidating}
Tero Karras, Miika Aittala, Timo Aila, and Samuli Laine.
\newblock Elucidating the design space of diffusion-based generative models.
\newblock In \emph{Conference on Neural Information Processing Systems (NeurIPS)}, 2022.

\bibitem[Karras et~al.(2024)Karras, Aittala, Lehtinen, Hellsten, Aila, and Laine]{karras2023analyzing}
Tero Karras, Miika Aittala, Jaakko Lehtinen, Janne Hellsten, Timo Aila, and Samuli Laine.
\newblock Analyzing and improving the training dynamics of diffusion models.
\newblock In \emph{IEEE Conference on Computer Vision and Pattern Recognition (CVPR)}, 2024.

\bibitem[Kawar et~al.(2023)Kawar, Zada, Lang, Tov, Chang, Dekel, Mosseri, and Irani]{kawar2023imagic}
Bahjat Kawar, Shiran Zada, Oran Lang, Omer Tov, Huiwen Chang, Tali Dekel, Inbar Mosseri, and Michal Irani.
\newblock Imagic: Text-based real image editing with diffusion models.
\newblock In \emph{IEEE Conference on Computer Vision and Pattern Recognition (CVPR)}, 2023.

\bibitem[kharchenkoirina()]{dancer_pose}
kharchenkoirina.
\newblock Fantasy fighting woman assassin actions in motion battle, hold daggers in hand.
\newblock \href{https://stock.adobe.com/images/fantasy-fighting-woman-assassin-actions-in-motion-battle-hold-daggers-in-hand-red-haired-girl-warrior-in-black-leather-costume-ninja-soldier-with-knives-red-long-hair-fluttering-fly-in-wind/483278018}{See source}.

\bibitem[Kim et~al.(2023)Kim, Lee, Kim, Ha, and Zhu]{kim2023dense}
Yunji Kim, Jiyoung Lee, Jin-Hwa Kim, Jung-Woo Ha, and Jun-Yan Zhu.
\newblock Dense text-to-image generation with attention modulation.
\newblock In \emph{IEEE International Conference on Computer Vision (ICCV)}, 2023.

\bibitem[Kirillov et~al.(2023)Kirillov, Mintun, Ravi, Mao, Rolland, Gustafson, Xiao, Whitehead, Berg, Lo, et~al.]{kirillov2023segment}
Alexander Kirillov, Eric Mintun, Nikhila Ravi, Hanzi Mao, Chloe Rolland, Laura Gustafson, Tete Xiao, Spencer Whitehead, Alexander~C Berg, Wan-Yen Lo, et~al.
\newblock Segment anything.
\newblock In \emph{IEEE International Conference on Computer Vision (ICCV)}, 2023.

\bibitem[Kolmogorov and Zabin(2004)]{kolmogorov2004energy}
Vladimir Kolmogorov and Ramin Zabin.
\newblock What energy functions can be minimized via graph cuts?
\newblock \emph{IEEE Transactions on Pattern Analysis and Machine Intelligence (TPAMI)}, 26\penalty0 (2), 2004.

\bibitem[Krakenimages.com()]{hand_source}
Krakenimages.com.
\newblock hand symbol.
\newblock \href{https://stock.adobe.com/images/hand-symbol/26799289}{See source}.

\bibitem[Kwatra et~al.(2003)Kwatra, Sch{\"o}dl, Essa, Turk, and Bobick]{kwatra2003graphcut}
Vivek Kwatra, Arno Sch{\"o}dl, Irfan Essa, Greg Turk, and Aaron Bobick.
\newblock Graphcut textures: Image and video synthesis using graph cuts.
\newblock \emph{ACM Transactions on Graphics (TOG)}, 22\penalty0 (3), 2003.

\bibitem[Lamontagne(2023)]{snake_source}
Gabrielle Lamontagne.
\newblock Snake.
\newblock \href{https://www.worldatlas.com/animals/snake.html}{See source}, 2023.

\bibitem[Lee et~al.(2023)Lee, Kim, Kim, and Sung]{lee2023syncdiffusion}
Yuseung Lee, Kunho Kim, Hyunjin Kim, and Minhyuk Sung.
\newblock Syncdiffusion: Coherent montage via synchronized joint diffusions.
\newblock In \emph{Neural Information Processing Systems (NeurIPS)}, 2023.

\bibitem[Li et~al.(2023)Li, Liu, Wu, Mu, Yang, Gao, Li, and Lee]{li2023gligen}
Yuheng Li, Haotian Liu, Qingyang Wu, Fangzhou Mu, Jianwei Yang, Jianfeng Gao, Chunyuan Li, and Yong~Jae Lee.
\newblock Gligen: Open-set grounded text-to-image generation.
\newblock In \emph{IEEE Conference on Computer Vision and Pattern Recognition (CVPR)}, 2023.

\bibitem[Liu et~al.(2022)Liu, Li, Du, Torralba, and Tenenbaum]{liu2022compositional}
Nan Liu, Shuang Li, Yilun Du, Antonio Torralba, and Joshua~B Tenenbaum.
\newblock Compositional visual generation with composable diffusion models.
\newblock In \emph{European Conference on Computer Vision (ECCV)}, 2022.

\bibitem[Lysenko.A()]{eye_source}
Lysenko.A.
\newblock Simple eye icon vector. eyesight pictogram in flat style.
\newblock \href{https://stock.adobe.com/images/simple-eye-icon-vector-eyesight-pictogram-in-flat-style/146119533}{See source}.

\bibitem[Ma et~al.(2024)Ma, Liang, Chen, and Lu]{ma2024subject}
Jian Ma, Junhao Liang, Chen Chen, and Haonan Lu.
\newblock Subject-diffusion: Open domain personalized text-to-image generation without test-time fine-tuning.
\newblock In \emph{ACM SIGGRAPH}, 2024.

\bibitem[martialred()]{musical_source}
martialred.
\newblock Quaver or eighth music / musical note flat icon for radio apps and websites.
\newblock \href{https://stock.adobe.com/images/quaver-or-eighth-music-musical-note-flat-icon-for-radio-apps-and-websites/117955609}{See source}.

\bibitem[Meng et~al.(2022)Meng, He, Song, Song, Wu, Zhu, and Ermon]{meng2021sdedit}
Chenlin Meng, Yutong He, Yang Song, Jiaming Song, Jiajun Wu, Jun-Yan Zhu, and Stefano Ermon.
\newblock Sdedit: Guided image synthesis and editing with stochastic differential equations.
\newblock In \emph{International Conference on Learning Representations (ICLR)}, 2022.

\bibitem[Mou et~al.(2024)Mou, Wang, Xie, Wu, Zhang, Qi, and Shan]{mou2024t2i}
Chong Mou, Xintao Wang, Liangbin Xie, Yanze Wu, Jian Zhang, Zhongang Qi, and Ying Shan.
\newblock T2i-adapter: Learning adapters to dig out more controllable ability for text-to-image diffusion models.
\newblock In \emph{Conference on Artificial Intelligence (AAAI)}, 2024.

\bibitem[Nichol et~al.(2022)Nichol, Dhariwal, Ramesh, Shyam, Mishkin, McGrew, Sutskever, and Chen]{nichol2021glide}
Alex Nichol, Prafulla Dhariwal, Aditya Ramesh, Pranav Shyam, Pamela Mishkin, Bob McGrew, Ilya Sutskever, and Mark Chen.
\newblock Glide: Towards photorealistic image generation and editing with text-guided diffusion models.
\newblock In \emph{International Conference on Machine Learning (ICML)}, 2022.

\bibitem[Oquab et~al.(2023)Oquab, Darcet, Moutakanni, Vo, Szafraniec, Khalidov, Fernandez, Haziza, Massa, El-Nouby, et~al.]{oquab2023dinov2}
Maxime Oquab, Timoth{\'e}e Darcet, Th{\'e}o Moutakanni, Huy Vo, Marc Szafraniec, Vasil Khalidov, Pierre Fernandez, Daniel Haziza, Francisco Massa, Alaaeldin El-Nouby, et~al.
\newblock Dinov2: Learning robust visual features without supervision.
\newblock In \emph{Transactions on Machine Learning Research (TMLR)}, 2023.

\bibitem[OZON()]{lamp_source}
AYDIN OZON.
\newblock Empty hour glass or sand watch.
\newblock \href{https://stock.adobe.com/images/empty-hour-glass-or-sand-watch/339236499}{See source}.

\bibitem[Parmar et~al.(2023)Parmar, Kumar~Singh, Zhang, Li, Lu, and Zhu]{parmar2023zero}
Gaurav Parmar, Krishna Kumar~Singh, Richard Zhang, Yijun Li, Jingwan Lu, and Jun-Yan Zhu.
\newblock Zero-shot image-to-image translation.
\newblock In \emph{ACM SIGGRAPH}, 2023.

\bibitem[Parmar et~al.(2024)Parmar, Park, Narasimhan, and Zhu]{parmar2024one}
Gaurav Parmar, Taesung Park, Srinivasa Narasimhan, and Jun-Yan Zhu.
\newblock One-step image translation with text-to-image models.
\newblock \emph{arXiv preprint arXiv:2403.12036}, 2024.

\bibitem[Patashnik et~al.(2023)Patashnik, Garibi, Azuri, Averbuch-Elor, and Cohen-Or]{patashnik2023localizing}
Or Patashnik, Daniel Garibi, Idan Azuri, Hadar Averbuch-Elor, and Daniel Cohen-Or.
\newblock Localizing object-level shape variations with text-to-image diffusion models.
\newblock In \emph{IEEE International Conference on Computer Vision (ICCV)}, 2023.

\bibitem[Peebles and Xie(2023)]{peebles2023scalable}
William Peebles and Saining Xie.
\newblock Scalable diffusion models with transformers.
\newblock In \emph{IEEE International Conference on Computer Vision (ICCV)}, 2023.

\bibitem[P{\'e}rez et~al.(2003)P{\'e}rez, Gangnet, and Blake]{perez2003poisson}
Patrick P{\'e}rez, Michel Gangnet, and Andrew Blake.
\newblock Poisson image editing.
\newblock In \emph{ACM SIGGRAPH}, 2003.

\bibitem[Phung et~al.(2024)Phung, Ge, and Huang]{phung2023grounded}
Quynh Phung, Songwei Ge, and Jia-Bin Huang.
\newblock Grounded text-to-image synthesis with attention refocusing.
\newblock In \emph{IEEE Conference on Computer Vision and Pattern Recognition (CVPR)}, 2024.

\bibitem[Rombach et~al.(2022)Rombach, Blattmann, Lorenz, Esser, and Ommer]{Rombach_2022_CVPR}
Robin Rombach, Andreas Blattmann, Dominik Lorenz, Patrick Esser, and Bj\"orn Ommer.
\newblock High-resolution image synthesis with latent diffusion models.
\newblock In \emph{IEEE Conference on Computer Vision and Pattern Recognition (CVPR)}, 2022.

\bibitem[Rother et~al.(2004)Rother, Kolmogorov, and Blake]{rother2004grabcut}
Carsten Rother, Vladimir Kolmogorov, and Andrew Blake.
\newblock ``grabcut'': interactive foreground extraction using iterated graph cuts.
\newblock \emph{ACM Transactions on Graphics (TOG)}, 23\penalty0 (3), 2004.

\bibitem[Rubinstein et~al.(2008)Rubinstein, Shamir, and Avidan]{rubinstein2008improved}
Michael Rubinstein, Ariel Shamir, and Shai Avidan.
\newblock Improved seam carving for video retargeting.
\newblock \emph{ACM Transactions on Graphics (TOG)}, 27\penalty0 (3), 2008.

\bibitem[Sarukkai et~al.(2024)Sarukkai, Li, Ma, R{\'e}, and Fatahalian]{sarukkai2024collage}
Vishnu Sarukkai, Linden Li, Arden Ma, Christopher R{\'e}, and Kayvon Fatahalian.
\newblock Collage diffusion.
\newblock In \emph{IEEE/CVF Winter Conference on Applications of Computer Vision (WACV)}, 2024.

\bibitem[Sauer et~al.(2023{\natexlab{a}})Sauer, Karras, Laine, Geiger, and Aila]{sauer2023stylegan}
Axel Sauer, Tero Karras, Samuli Laine, Andreas Geiger, and Timo Aila.
\newblock Stylegan-t: Unlocking the power of gans for fast large-scale text-to-image synthesis.
\newblock In \emph{International Conference on Machine Learning (ICML)}, 2023{\natexlab{a}}.

\bibitem[Sauer et~al.(2023{\natexlab{b}})Sauer, Lorenz, Blattmann, and Rombach]{sauer2023adversarial}
Axel Sauer, Dominik Lorenz, Andreas Blattmann, and Robin Rombach.
\newblock Adversarial diffusion distillation.
\newblock \emph{arXiv preprint arXiv:2311.17042}, 2023{\natexlab{b}}.

\bibitem[Shirakawa and Uchida(2024)]{shirakawa2024noisecollage}
Takahiro Shirakawa and Seiichi Uchida.
\newblock Noisecollage: A layout-aware text-to-image diffusion model based on noise cropping and merging.
\newblock In \emph{IEEE Conference on Computer Vision and Pattern Recognition (CVPR)}, 2024.

\bibitem[Simonyan and Zisserman(2015)]{simonyan2015vgg}
Karen Simonyan and Andrew Zisserman.
\newblock Very deep convolutional networks for large-scale image recognition.
\newblock In \emph{International Conference on Learning Representations (ICLR)}, 2015.

\bibitem[Sobek()]{waffle_source}
Werner Sobek.
\newblock Heydar aliyev center.
\newblock \href{https://www.wernersobek.com/focus/hac/}{See source}.

\bibitem[Sohl-Dickstein et~al.(2015)Sohl-Dickstein, Weiss, Maheswaranathan, and Ganguli]{sohl2015deep}
Jascha Sohl-Dickstein, Eric Weiss, Niru Maheswaranathan, and Surya Ganguli.
\newblock Deep unsupervised learning using nonequilibrium thermodynamics.
\newblock In \emph{International Conference on Machine Learning (ICML)}, 2015.

\bibitem[Song et~al.(2021)Song, Sohl-Dickstein, Kingma, Kumar, Ermon, and Poole]{song2020score}
Yang Song, Jascha Sohl-Dickstein, Diederik~P Kingma, Abhishek Kumar, Stefano Ermon, and Ben Poole.
\newblock Score-based generative modeling through stochastic differential equations.
\newblock In \emph{International Conference on Learning Representations (ICLR)}, 2021.

\bibitem[Song et~al.(2023)Song, Zhang, Lin, Cohen, Price, Zhang, Kim, and Aliaga]{objectstitch}
Yizhi Song, Zhifei Zhang, Zhe Lin, Scott Cohen, Brian Price, Jianming Zhang, Soo~Ye Kim, and Daniel Aliaga.
\newblock Objectstitch: Object compositing with diffusion model.
\newblock In \emph{IEEE Conference on Computer Vision and Pattern Recognition (CVPR)}, 2023.

\bibitem[Szeliski et~al.(2011)Szeliski, Uyttendaele, and Steedly]{szeliski2011fast}
Richard Szeliski, Matthew Uyttendaele, and Drew Steedly.
\newblock Fast poisson blending using multi-splines.
\newblock In \emph{IEEE International Conference on Computational Photography (ICCP)}, 2011.

\bibitem[Tao et~al.(2022)Tao, Tang, Wu, Sebe, Jing, Wu, and Bao]{tao2020df}
Ming Tao, Hao Tang, Songsong Wu, Nicu Sebe, Xiao-Yuan Jing, Fei Wu, and Bingkun Bao.
\newblock Df-gan: Deep fusion generative adversarial networks for text-to-image synthesis.
\newblock In \emph{IEEE Conference on Computer Vision and Pattern Recognition (CVPR)}, 2022.

\bibitem[Tewel et~al.(2024)Tewel, Kaduri, Gal, Kasten, Wolf, Chechik, and Atzmon]{tewel2024training}
Yoad Tewel, Omri Kaduri, Rinon Gal, Yoni Kasten, Lior Wolf, Gal Chechik, and Yuval Atzmon.
\newblock Training-free consistent text-to-image generation.
\newblock \emph{ACM Transactions on Graphics (TOG)}, 43\penalty0 (4), 2024.

\bibitem[Tumanyan et~al.(2023)Tumanyan, Geyer, Bagon, and Dekel]{tumanyan2022plugandplay}
Narek Tumanyan, Michal Geyer, Shai Bagon, and Tali Dekel.
\newblock Plug-and-play diffusion features for text-driven image-to-image translation.
\newblock In \emph{IEEE Conference on Computer Vision and Pattern Recognition (CVPR)}, 2023.

\bibitem[Vogiatzis et~al.(2005)Vogiatzis, Torr, and Cipolla]{vogiatzis2005multi}
George Vogiatzis, Philip~HS Torr, and Roberto Cipolla.
\newblock Multi-view stereo via volumetric graph-cuts.
\newblock In \emph{IEEE Conference on Computer Vision and Pattern Recognition (CVPR)}, 2005.

\bibitem[Wang et~al.(2004)Wang, Bovik, Sheikh, and Simoncelli]{wang2004image}
Zhou Wang, Alan~C Bovik, Hamid~R Sheikh, and Eero~P Simoncelli.
\newblock Image quality assessment: from error visibility to structural similarity.
\newblock \emph{IEEE transactions on image processing}, 13\penalty0 (4):\penalty0 600--612, 2004.

\bibitem[Wiiii(2008)]{temple_source}
Wiiii.
\newblock Tōdai-ji kon-dō, at nara japan.
\newblock \href{https://en.wikipedia.org/wiki/T\%C5\%8Ddai-ji\#/media/File:T\%C5\%8Ddai-ji_Kon-d\%C5\%8D.jpg}{See source}, 2008.

\bibitem[Wu et~al.(2019)Wu, Zheng, Zhang, and Huang]{wu2019gp}
Huikai Wu, Shuai Zheng, Junge Zhang, and Kaiqi Huang.
\newblock Gp-gan: Towards realistic high-resolution image blending.
\newblock In \emph{ACM Multimedia (MM)}, 2019.

\bibitem[Yang et~al.(2023)Yang, Gu, Zhang, Zhang, Chen, Sun, Chen, and Wen]{yang2023paint}
Binxin Yang, Shuyang Gu, Bo Zhang, Ting Zhang, Xuejin Chen, Xiaoyan Sun, Dong Chen, and Fang Wen.
\newblock Paint by example: Exemplar-based image editing with diffusion models.
\newblock In \emph{IEEE Conference on Computer Vision and Pattern Recognition (CVPR)}, 2023.

\bibitem[Ye et~al.(2023)Ye, Zhang, Liu, Han, and Yang]{ye2023ip}
Hu Ye, Jun Zhang, Sibo Liu, Xiao Han, and Wei Yang.
\newblock Ip-adapter: Text compatible image prompt adapter for text-to-image diffusion models.
\newblock \emph{arXiv preprint arXiv:2308.06721}, 2023.

\bibitem[Yu et~al.(2022)Yu, Xu, Koh, Luong, Baid, Wang, Vasudevan, Ku, Yang, Ayan, et~al.]{yu2022scaling}
Jiahui Yu, Yuanzhong Xu, Jing~Yu Koh, Thang Luong, Gunjan Baid, Zirui Wang, Vijay Vasudevan, Alexander Ku, Yinfei Yang, Burcu~Karagol Ayan, et~al.
\newblock Scaling autoregressive models for content-rich text-to-image generation.
\newblock In \emph{International Conference on Machine Learning (ICML)}, 2022.

\bibitem[Zhang et~al.(2020)Zhang, Wen, and Shi]{zhang2020deep}
Lingzhi Zhang, Tarmily Wen, and Jianbo Shi.
\newblock Deep image blending.
\newblock In \emph{IEEE/CVF Winter Conference on Applications of Computer Vision (WACV)}, 2020.

\bibitem[Zhang et~al.(2023{\natexlab{a}})Zhang, Rao, and Agrawala]{zhang2023adding}
Lvmin Zhang, Anyi Rao, and Maneesh Agrawala.
\newblock Adding conditional control to text-to-image diffusion models.
\newblock In \emph{IEEE International Conference on Computer Vision (ICCV)}, 2023{\natexlab{a}}.

\bibitem[Zhang et~al.(2018)Zhang, Isola, Efros, Shechtman, and Wang]{zhang2018perceptual}
Richard Zhang, Phillip Isola, Alexei~A Efros, Eli Shechtman, and Oliver Wang.
\newblock The unreasonable effectiveness of deep features as a perceptual metric.
\newblock In \emph{IEEE Conference on Computer Vision and Pattern Recognition (CVPR)}, 2018.

\bibitem[Zhang et~al.(2023{\natexlab{b}})Zhang, Dong, Tang, Huang, Huang, Ma, Lee, Deussen, and Xu]{zhang2023prospect}
Yuxin Zhang, Weiming Dong, Fan Tang, Nisha Huang, Haibin Huang, Chongyang Ma, Tong-Yee Lee, Oliver Deussen, and Changsheng Xu.
\newblock Prospect: Prompt spectrum for attribute-aware personalization of diffusion models.
\newblock \emph{ACM Transactions on Graphics (TOG)}, 42\penalty0 (6), 2023{\natexlab{b}}.

\bibitem[Zhao et~al.(2023)Zhao, Chen, Chen, Bao, Hao, Yuan, and Wong]{zhao2024uni}
Shihao Zhao, Dongdong Chen, Yen-Chun Chen, Jianmin Bao, Shaozhe Hao, Lu Yuan, and Kwan-Yee~K. Wong.
\newblock Uni-controlnet: All-in-one control to text-to-image diffusion models.
\newblock In \emph{Neural Information Processing Systems (NeurIPS)}, 2023.

\bibitem[Zheng et~al.(2023)Zheng, Zhou, Li, Qi, Shan, and Li]{Zheng_2023_CVPR}
Guangcong Zheng, Xianpan Zhou, Xuewei Li, Zhongang Qi, Ying Shan, and Xi Li.
\newblock Layoutdiffusion: Controllable diffusion model for layout-to-image generation.
\newblock In \emph{IEEE Conference on Computer Vision and Pattern Recognition (CVPR)}, 2023.

\bibitem[Zhu et~al.(2019)Zhu, Pan, Chen, and Yang]{zhu2019dm}
Minfeng Zhu, Pingbo Pan, Wei Chen, and Yi Yang.
\newblock Dm-gan: Dynamic memory generative adversarial networks for text-to-image synthesis.
\newblock In \emph{IEEE Conference on Computer Vision and Pattern Recognition (CVPR)}, 2019.

\end{thebibliography}
}

\clearpage
\appendix
\section*{Appendix}
\label{sec:app}

In Appendix \ref{sec:app_runtime}, we discuss runtime and storage performance of our method. Then, we show more results (Appendix  \ref{sec:app_results}), qualitative comparisons with baselines (Appendix \ref{sec:app_baselines}), and additional ablations (Appendix \ref{sec:app_ablations}). Next, we provide implementation details of our baselines (Appendix \ref{sec:app_baselines_details}), the user survey details (Appendix \ref{sec:app_user_survey}), and result details (Appendix \ref{sec:app_result_details}). Finally, we discuss our work's limitations in Appendix \ref{sec:app_limitations}.

\section{Runtime and Storage}
\label{sec:app_runtime}

Our graph cut runs in ${\sim}1$ sec in total, which includes preprocessing (e.g., PCA space computation, computing energy costs) and the actual graph-cut optimization. In theory, the graph-cut optimization depends on the number of variables (feature map resolution) and candidate labels (number of images). In practice, however, there is very little overhead. For our results ($\sim$64x64 feature maps with 2-5 images), the solver takes $\sim$3ms on average to find the solution. 

The blending stage takes ${\sim}3$ seconds on an NVIDIA A6000 GPU.
For reference, one forward pass of vanilla ControlNet takes ${\sim}2$ seconds on the same GPU.
Storage space for the initially generated $QKV$ features depends on image resolution and is  ${\sim}2$GB for a $512\times512$ image. 

While we resorted to storing the features on disk, most modern GPUs have enough VRAM to generate multiple images in a single batch, obviating the need to store features on disk. For example, A6000 (48GB) can generate up to $60$ images for SD1.5 ControlNet.  
Hence, one can re-generate the image stack and its QKV features in the same batch as the composite image during the second pass. Alternatively, one could store a subset of features with a trade-off of lower appearance fidelity (please see Appendix \ref{sec:app_ablation_injection} for more details).

\section{Graph Cut Parameters}
\begin{figure}
  \includegraphics[width=\linewidth]{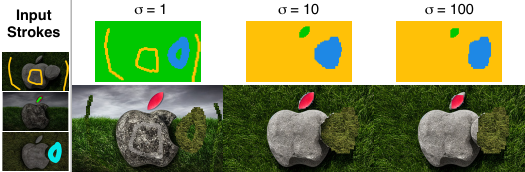}
  \vspace{-18pt}
  \caption{{\bf Graph Cut: $\sigma$ parameter}. A large $\sigma$ 
prevents cuts from aligning with semantic boundaries, and a small $\sigma$ is overly sensitive to feature differences, causing cuts to fall directly on the strokes. Our graph cut parameter ($\sigma=10$) strikes a balance and segments image regions along semantic boundaries.}
  \label{fig:graphcut_hyperparam}
  \vspace{-10pt}
\end{figure}

We use the same graph cut parameters, C = $10^6, \lambda = 100, \sigma = 10$, across all our results. $\sigma$ controls how quickly the pairwise cost decreases with feature changes; a large $\sigma$ 
prevents cuts from aligning with semantic boundaries, and a small $\sigma$ is overly sensitive to feature differences, causing cuts to fall directly on the strokes (Figure \ref{fig:graphcut_hyperparam}). 
$\lambda$ scales the pairwise costs to reduce rounding errors for the solver, and $C=10^6$ ensures unary costs dominate, enforcing a hard constraint for strokes.

\section{Additional Results} 
\label{sec:app_results}

\begin{figure*}[h]
\centering
  \includegraphics[width=\linewidth]{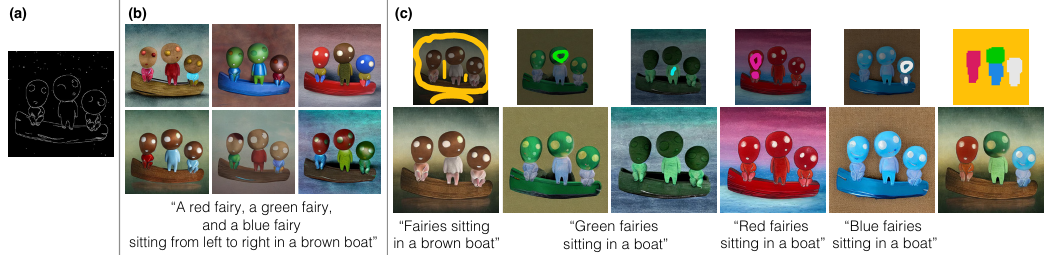}
  \caption{\textbf{Result: Prompt Alignment}. (a) ControlNet input condition. (b) Vanilla ControlNet struggles to adhere to the long, complicated prompt. (c) With Generative Photomontage, the user can instead generate a stack of images with multiple, short prompts, where each spatial region has at least one correct image within the stack. Generative Photomontage composites the user-selected regions together, where each scene element has the correct color according to the original, long prompt.}
  \label{fig:app_fairies}
\end{figure*}

\begin{figure*}[h]
\centering
  \includegraphics[width=\linewidth]{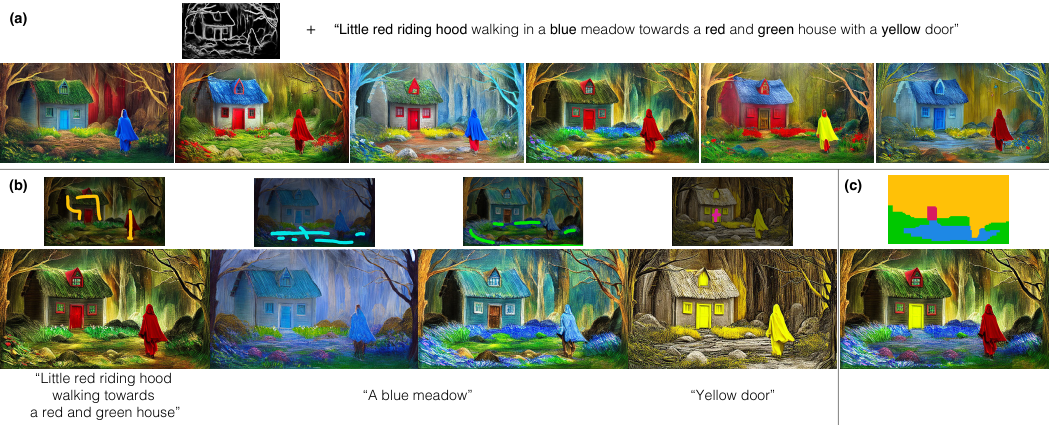}
  \caption{\textbf{Result: Prompt Alignment}. Another example of using Generative Photomontage to increase prompt alignment. (a) Vanilla ControlNet struggles to adhere to the long, complicated prompt, i.e., it assigns the wrong color to scene elements. (b) The user can instead generate a stack of images with multiple, short prompts, where each spatial region has at least one correct image within the stack. User strokes are shown on top of each image. (c) Generative Photomontage composites the user-selected regions together (bottom), where each scene element has the correct color according to the original, long prompt. Top: feature-space graph-cut result.}
  \label{fig:red}
\end{figure*}

\begin{figure*}[h]
\centering
  \includegraphics[width=\linewidth]{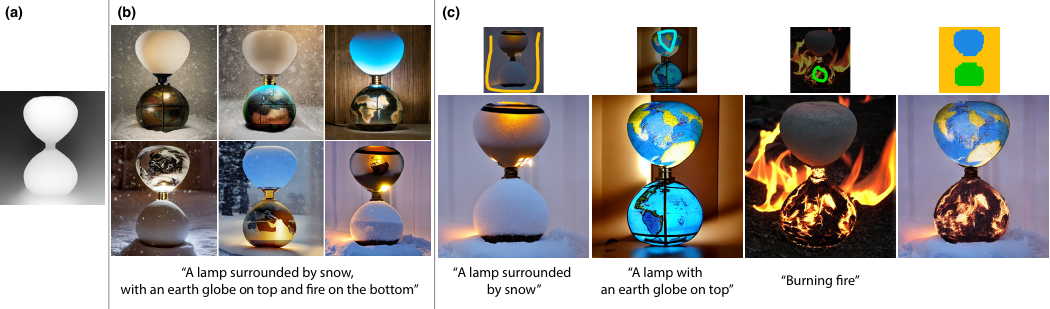}
  \caption{\textbf{Result: Prompt Alignment}. (a) ControlNet input condition. (b) Vanilla ControlNet struggles to follow all aspects of the long prompt. (c) The user can instead break up the prompt into multiple, shorter prompts, and use our method to composite the outputs to create the desired image.}
  \label{fig:lamp}
\end{figure*}

\begin{figure*}[h]
\centering
  \includegraphics[width=\linewidth]{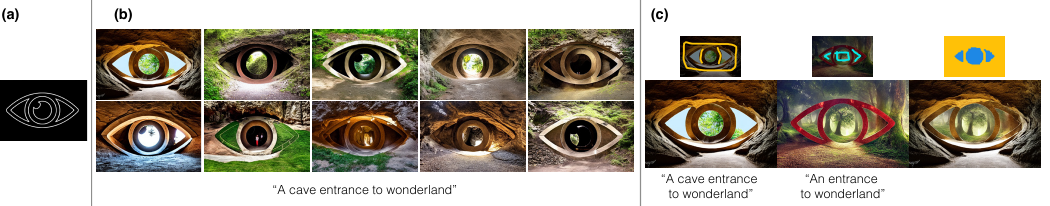}
  \caption{\textbf{Result: Prompt Alignment}. (a) ControlNet input condition. (b) Due to the unconventional shape and prompt combination (i.e., cave, eye shape, and wonderland), vanilla ControlNet struggles to adhere to all aspects of the prompt. (c) Using Generative Photomontage, users can generate multiple results with varying complexity of prompts, and composite the results to form the desired image.}
  \label{fig:eye}
\end{figure*}

We include our full results in Figures \ref{fig:app_appearance1}-\ref{fig:eye}. We show applications of appearance mixing (Figures \ref{fig:app_appearance1} and \ref{fig:app_appearance2}), shape and artifacts correction (Figure \ref{fig:app_shapes}), and prompt alignment (Figures \ref{fig:app_fairies}, \ref{fig:red}, \ref{fig:lamp} and \ref{fig:eye}).

\section{Additional Baseline Qualitative Comparisons}
\label{sec:app_baselines}
\begin{figure*}[h]
  \includegraphics[width=\linewidth]{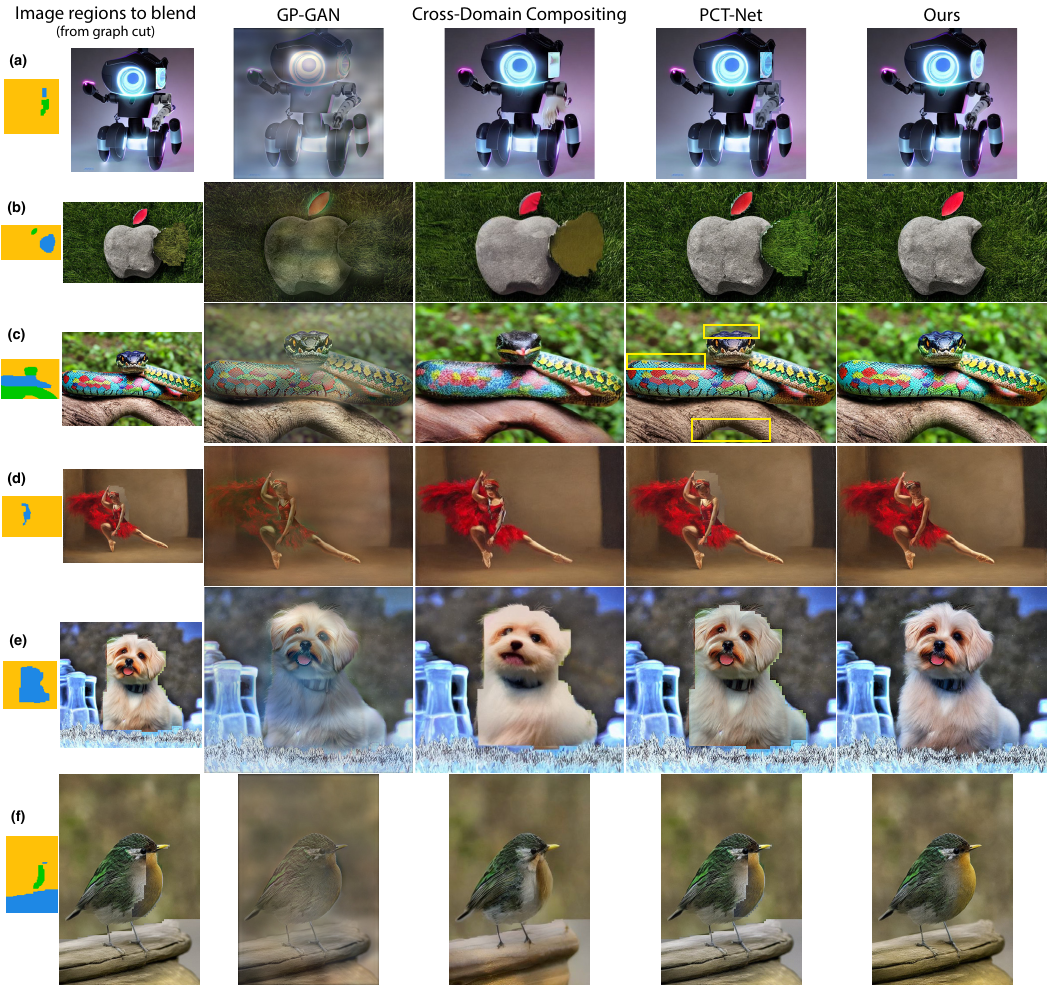}
  \caption{\textbf{Qualitative Comparison (Baselines)}. Leftmost column: Image regions to blend (output of graph-cut optimization). Graph-cut in diffusion feature space is visualized on the left, and the image-space composite of that graph-cut is visualized on the right. GP-GAN \cite{wu2019gp} does not preserve local appearances and tends to smooth out color. Cross-Domain Compositing \cite{hachnochi2023crossdomain} may struggle to blend regions harmoniously, particularly in non-oil painting style images (a-c, e-f). PCT-Net \cite{pctnet} changes the interior color of local regions but struggles to blend away the seams.}
  \label{fig:supp_qualitative2}
\end{figure*}

We show additional qualitative comparisons of our method with baselines in Figures \ref{fig:supp_qualitative_full}, \ref{fig:supp_qualitative}, and \ref{fig:supp_qualitative2}. Our method outperforms the baselines in terms of preserving local appearances and blending harmoniously.

\section{Additional Ablations}
\label{sec:app_ablations}

Here, we show additional ablation results on our graph-cut segmentation, the features used for the graph cut, and the self-attention feature injection. 

\subsection{Multi-Image Segmentation}
In Table \ref{table:ablation_segmentation}, we show quantitative results of ablating our graph-cut segmentation with SAM \cite{kirillov2023segment} and IDP's pixel-space graph cut \cite{agarwala2004interactive}. In both cases, the masked LPIPS and PSNR scores became worse, reflecting a lower fidelity of local appearances due to less accurate segmentations. Please see Section \ref{sec:multi_img_seg} and Fig. \ref{fig:dp_ours} and \ref{fig:sam} for qualitative examples.

\begin{table}[!t]
\centering
\resizebox{\linewidth}{!}{
\renewcommand{\arraystretch}{0.8}
\setlength{\tabcolsep}{5pt}
\begin{tabular}{ l c c c c} 
\toprule
& \shortstack[c]{\textbf{Masked} \\ \textbf{LPIPS}}$\downarrow$ & 
\shortstack[c]{\textbf{Masked} \\ \textbf{SSIM}}$\uparrow$ & 
\textbf{PSNR} $\uparrow$ & 
\shortstack[c]{\textbf{Seam Gradient Score} \\
\footnotesize{min, avg, max} \\
\footnotesize{0.255, 0.340, 0.425} 
} \\
\midrule
 \textbf{Ours} & \textbf{0.123} &  \textbf{0.815} & \textbf{22.46} & 0.339\\
 w/ SAM & 0.223 & 0.697 & 17.71 & 0.325 \\
 \shortstack[l]{w/ IDP graph cut} & 0.146 & 0.786 & 21.28 &  0.312 \\
\bottomrule 
\end{tabular}
}
\caption{{\bf Ablation: Segmentation}.
}
\label{table:ablation_segmentation}
\vspace{-10pt}
\end{table}

\subsection{Graph Cut}
\label{sec:app_gc_ablations}
\begin{figure*}
\centering
  \includegraphics[width=\linewidth]{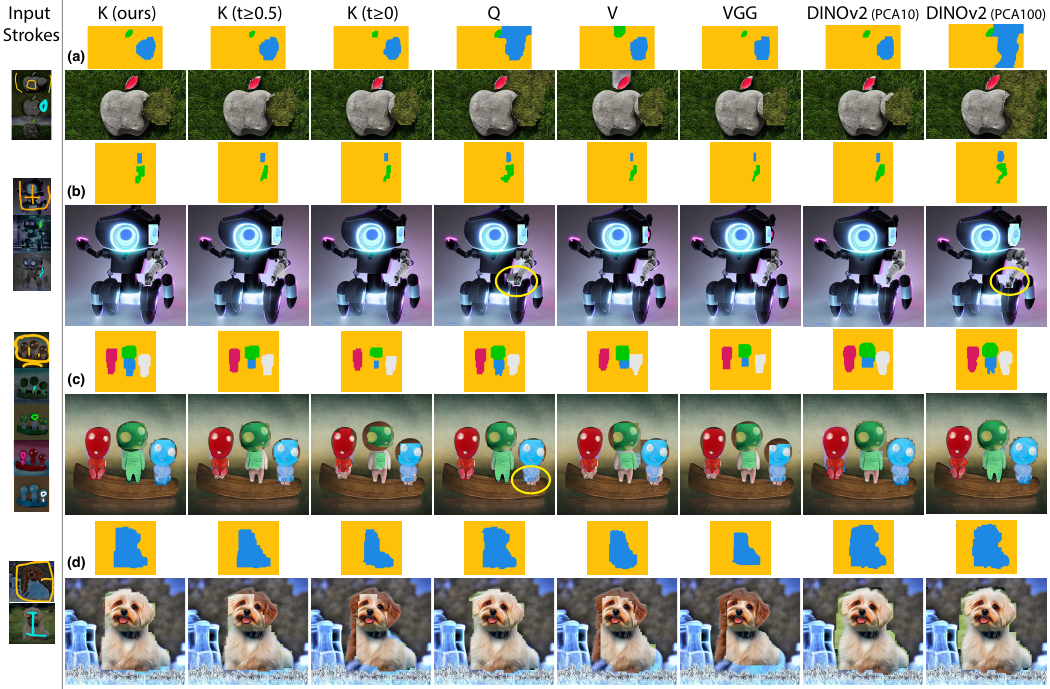}
  \caption{\textbf{Ablation: Graph Cut Features}. Our method uses the self-attention $K$ features to compute pairwise seam costs in the optimization. Here, we experiment with using the $K$ features averaged across different time steps, the other self-attention features $Q$ and $V$, as well as VGG \cite{simonyan2015vgg} and DINOv2 \cite{oquab2023dinov2} features. $K$ features performs the best, with $Q$ features a close second.}
  \label{fig:ablation_gc}
  \vspace{-10pt}
\end{figure*}

Our method uses the output features of the key projection matrix from self-attention layers ($K$ features) to compute the graph cut energy terms, i.e., the pairwise terms that dictate seam costs. Here, we consider alternative features, as discussed in Section \ref{sec:multi_img_seg}, and show its results in Figure \ref{fig:ablation_gc}. $K$ features performs the best, with $Q$ features a close second. Using $K$ features at the last timestep is also better than averaging $K$ features across earlier timesteps.

\myparagraph{$K$ features timesteps.} In our method, we use the $K$ features from the last denoising step to set up the seam costs, when the image mostly formed. Here, we experiment with using earlier timesteps. Specifically, we experiment with $K$ features averaged over: 1) the last half of denoising steps ($t\geq0.5$), and 2) all time steps ($t\geq0$). As shown in Figure \ref{fig:ablation_gc}, incorporating $K$ features from earlier time steps lead to under-segmentations, where objects' boundaries are not captured. This aligns with observations from previous works \cite{cao2023masactrl, zhang2023prospect} that earlier time steps form content and layout, while later time steps refine detailed appearance. By averaging earlier time steps, we weigh low-frequency content more heavily, thus leading to under-segmentations near boundaries.

\myparagraph{$Q$ and $V$ features.} Here, we use output features from query and value projection matrices, namely $Q$ and $V$ features, for segmentation. $Q$ features usually lead to similar segmentations as $K$ features but may exhibit over-and under-segmentations near boundaries in some cases (circled in Figure \ref{fig:ablation_gc}). $V$ features generally do not lead to seams that align with semantic boundaries.

\myparagraph{VGG Features.} We experimented with VGG features \cite{simonyan2015vgg} instead of diffusion features. Specifically, we extract the features from the second block's ReLU layer, following Johnson et al.~\shortcite{johnson2016perceptual}. Figure \ref{fig:ablation_gc} shows that VGG features typically leads to under-segmentations.

\myparagraph{DINOv2 Features.} We also experimented with DINOv2 features \cite{oquab2023dinov2} in place of diffusion features. Specifically, we used the small DINOv2 model, which outputs features per $14\times14$ patch. For a fair comparison, we resized all images such that the size of output features would equal that of $K$ features. As shown in Figure \ref{fig:ablation_gc}, if we directly apply our graph cut method (which uses the features' top-10 PCA components), we see errors in segmentation. If we use the top-100 PCA components, we get better segmentations for most examples but not all (e.g., the dog in Figure \ref{fig:ablation_gc}d).

\begin{figure*}
    \centering
        \includegraphics[width=0.6\linewidth]{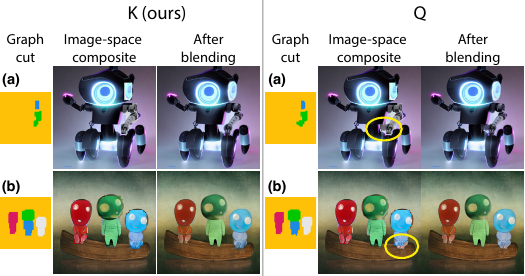}
      \caption{\textbf{Effects of Segmentation on Feature Blending}. Our feature blending method is robust to small errors in segmentation boundaries. For example,  while $Q$ features may result in over- or under-segmentations near boundaries, our feature blending method blends the seams well, such that the difference in the resulting images is not too noticeable.}
      \label{fig:ablation_gc2}
\end{figure*}

\myparagraph{Discussion.} While we chose to go with $K$ features due to the best segmentation, we found that our feature blending method is robust to small errors in segmentation boundaries. For example, the segmentations we get from $Q$ features may result in over- or under-segmentations near boundaries, but our feature blending method blends the seams well, such that the difference in the resulting images is not too noticeable (Figure \ref{fig:ablation_gc2}).

\subsection{Feature Injection}
\label{sec:app_ablation_injection}

\begin{figure}
  \includegraphics[width=\linewidth]{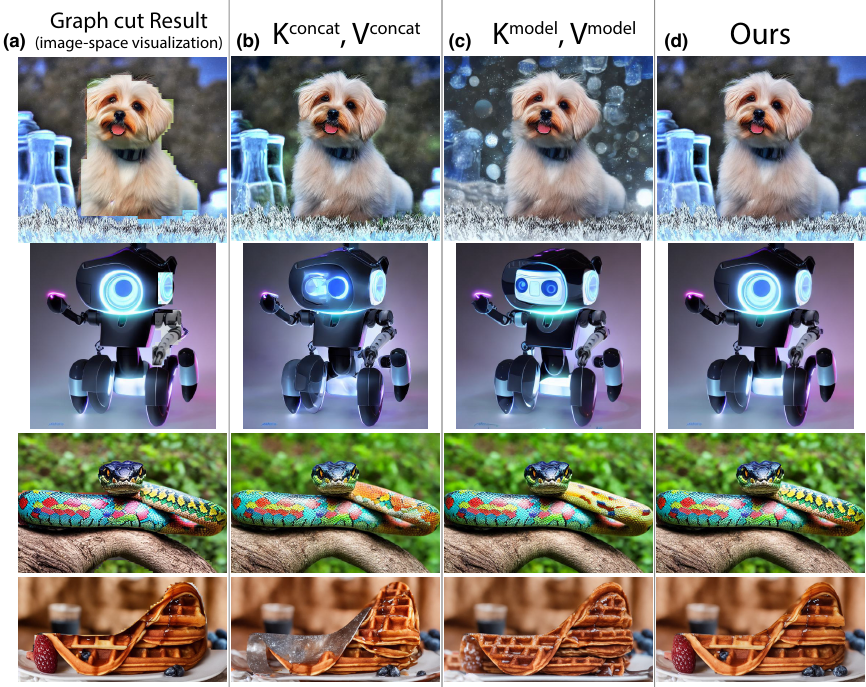}
  \vspace{-18pt}
  \caption{{\bf Ablation with Alternative Self-Attention Injection Strategies}. (a) Input graph-cut segmentation, visualized here by resizing the diffusion-feature masks to match image dimensions and compositing in pixel space. (b) Inspired by StyleAligned \cite{hertz2024style}, we replace $K^{\text{comp}}, V^{\text{comp}}$ with $K^{\text{concat}} = [K^1, K^2, ...,K^N]$ and $V^{\text{concat}} = [V^1, V^2, ...,V^N]$. (c) Similar to Equation 4, we injected $K^{\text{model}}$ and $V^{\text{model}}$ for the base image. (d) Our method can preserve local image appearances while harmoniously blending them. Please zoom in for details.}
  \label{fig:ablation}
  \vspace{-10pt}
\end{figure}

In Sec \ref{sec:ablation_main}, we showed quantitative results of (1) using shared (concatenated) key and value features, i.e, $K^{\text{concat}} =[K^1, K^2, ...,K^N]$ and $V^{\text{concat}} = [V^1, V^2, ...,V^N]$, inspired by Hertz et al.~\shortcite{hertz2024style}; and (2) using $K^{\text{model}}$ and $V^{\text{model}}$ for the base image, similar to Equation \ref{eq:q_comp}. Here, Figure \ref{fig:ablation} shows qualitative samples of final outputs along with the intermediate feature-based graph cut, visualized in image space. 

As shown, the alternative injection schemes can change the local image appearance, such as the dog's background, the robot's head, the snake color, and the waffle. 
Our results align with the observation \cite{alaluf2023cross, cao2023masactrl} that $Q$ features influence the image structure, and $K$ and $V$ features influence the image appearance. 

$K^{\text{concat}}, V^{\text{concat}}$ allows $Q$ features to match with keys that are not from the target image region and can thus result in appearance changes. Our method only makes the target image region's keys and values available and, hence, is able to preserve local appearances. 
Using $K^{\text{model}}, V^{\text{model}}$ for the base image also allows appearances to change. By injecting $K^B$ and $V^B$ of the base image stored during initial generation, our method can preserve their appearance during blending.

Our method injects the composite $Q^{\text{comp}}$, $K^{\text{comp}}$, $V^{\text{comp}}$ into the U-Net's self-attention maps for all layers and time steps. Following related works in self-attention injection \cite{cao2023masactrl, alaluf2023cross}, we also consider injecting only in decoder layers and in later time steps.

\begin{figure*}
    \centering
  \includegraphics[width=1.0\linewidth]{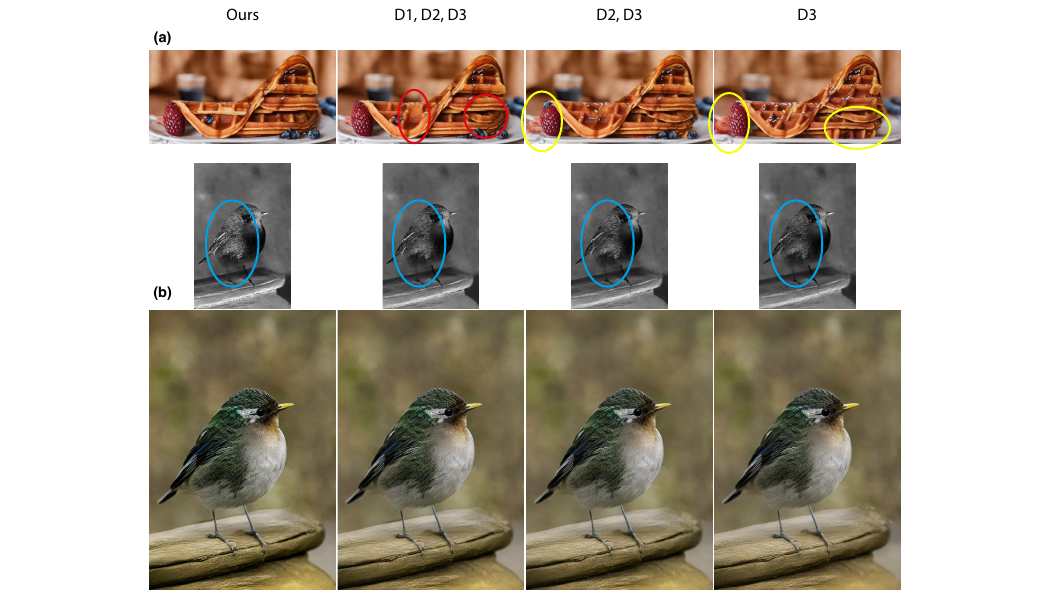}
  \caption{\textbf{Ablation: Injection Layers}. Experiments of injecting $Q^{\text{comp}}$, $K^{\text{comp}}$, $V^{\text{comp}}$ only in decoder blocks (D1, D2, D3) of ControlNet. (a) Injecting only in the decoder blocks leads to changes in the waffle interior (circled in red). As we reduce the number of injected layers, more artifacts and local changes appear, such as the extra strawberry and missing blueberries (circled in yellow). (b) Injecting only in the decoder layers also tends to reduce color vibrancy in the composite image. Saturation is visualized above each image (white: high saturation; black: low saturation). As shown, the color saturation of bird feathers is reduced in the ablated results (circled in blue).}
  \label{fig:ablation_layers}
\end{figure*}

\myparagraph{Layers.} ControlNet with Stable Diffusion v1.5 has three decoder blocks (D1, D2, D3). We experimented with injecting in subsets of these blocks (D1-D3, D2-D3, D3). 
In many cases, injecting only the decoder layers (D1-D3) lead to similar results as our method (which injects all layers). However, in some cases, it may lead to artifacts and structural changes (e.g., waffle texture in Figure \ref{fig:ablation_layers}a), and it often reduces color vibrancy and saturation (e.g., bird feathers in Figure \ref{fig:ablation_layers}b) in the composite output.

\begin{figure*}
    \centering
  \includegraphics[width=1.0\linewidth]{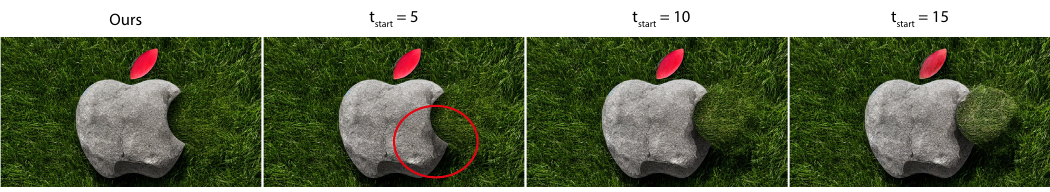}
  \caption{\textbf{Ablation: Injection Time Steps}. Experiments of injecting $Q^{\text{comp}}$, $K^{\text{comp}}$, $V^{\text{comp}}$ after a number of time steps ($t_{\text{start}}$) during the denoising process (total: $20$ time steps). As shown, starting the injection later leads to more artifacts. Our results ($t_{\text{start}} = 0$) completely remove the extra rock at the apple bite and also remove the shadow of the extra rock. However, when $t_{\text{start}} = 5$, we see the shadow still remains on the base rock. This aligns with previous observations that earlier time steps form image layout and shape \cite{zhang2023prospect, cao2023masactrl}, so starting the injection process later reduces the model's ability to adapt image structure near the seams.}
  \label{fig:ablation_timesteps}
\end{figure*}

\myparagraph{Time Steps.} We used a total of $20$ denoising time steps for all our results. Prior work shows that earlier denoising time steps form image layout and shape, while later time steps form detailed appearance \cite{zhang2023prospect, cao2023masactrl}. We experimented with applying our injection after $5$, $10$, and $15$ time steps.  As shown in Figure \ref{fig:ablation_timesteps}, starting the injection later tends to create artifacts in the blended output.
In most cases, starting the injection after $5$ time steps leads to comparable results to our method (which injects all time steps). However, in scenes that require the image structure to adapt near the seams, doing so may prevent the model from adapting sufficiently. For example, if we inject after $5$ time steps, the shadow from the removed rock at the apple bite still shows up in the final image (circled in red), whereas it is completely removed if we apply the injection for all time steps.

\myparagraph{Discussion.} If the required adaption near seams is minor, and the user does not mind changes to local appearances, one can consider applying the injection only in decoder layers (D1-D3) and after $5$ time steps. 
However, if the user wishes to maximize adaptation near seams and to maximize local appearance fidelity, we recommend using our full injection approach.

\section{Baseline Implementation Details}
\label{sec:app_baselines_details}
Below, we describe the implementation details of the baselines.

\myparagraph{Interactive Digital Photomontage \cite{agarwala2004interactive}.} 
For their pixel-space graph cut, we used their ``match edge'' seam objective, where the pairwise term $E(p, q, L_p, L_q)$ is computed based on a Sobel filter on the RGB values. As shown in Figure \ref{fig:dp_ours}, the edge strength computed from the filter may be sensitive to low-level changes in color, which could cause their graph cut to select seams that fall on undesired edges. Diffusion feature-based graph cut, on the other hand, selects seams that are more aligned with semantic features. 

\begin{figure*}[h]
  \includegraphics[width=\linewidth]{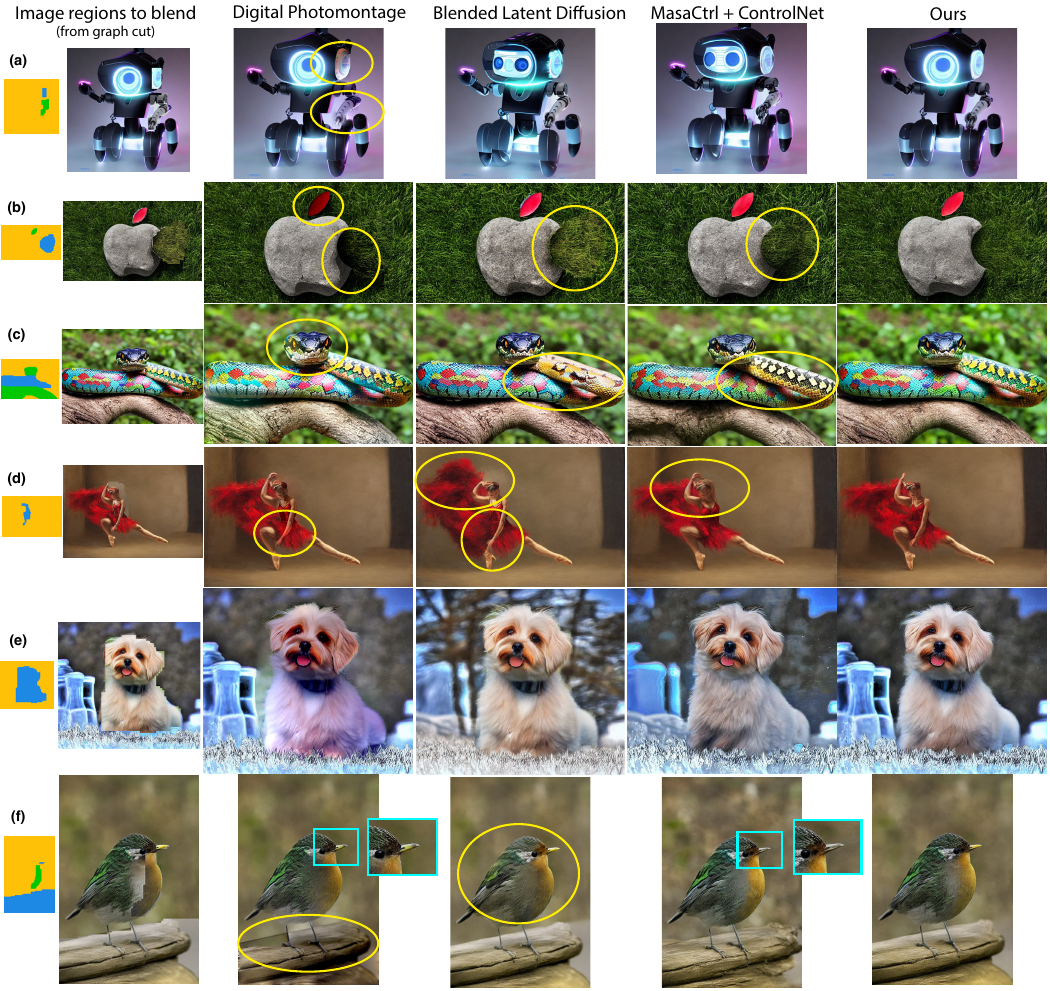}
  \caption{\textbf{Qualitative Comparison (Baselines)}. Leftmost column: Image regions to blend (output of graph-cut optimization). Graph-cut in diffusion feature space is visualized on the left, and the image-space 
  composite of that graph-cut is visualized on the right. 
  Interactive Digital Photomontage \cite{agarwala2004interactive}: pixel-space graph-cut may cause seams to fall on undesired edges (see Figure \ref{fig:dp_ours} also), and their gradient-domain blending often fails to preserve color, e.g., the bird's yellow beak is not preserved in (f). Blended latent diffusion \cite{avrahami2023blended} and MasaCtrl+ControlNet \cite{cao2023masactrl} may lead to color changes (c, f) and structure changes (a, b, d, e).}
  \label{fig:supp_qualitative_full}
\end{figure*}

After their pixel-space graph cut, we followed their approach and used Poisson blending~\cite{perez2003poisson} as a post-process for smoothing composite regions together.
Because generated images~\cite{perez2003poisson} tend to have wide color variations, gradient-domain blending often leads to changes in color, such as the marked regions in Figure \ref{fig:supp_qualitative_full}. Our method is better at preserving local color and image appearances.

\myparagraph{Blended Latent Diffusion \cite{avrahami2023blended}.} BLD blends images by combining their noise at each diffusion step. For this baseline, we inject the noises of image $i=2...N$ into the noise of the base image $i=1$. 
As shown in Figure \ref{fig:supp_qualitative_full}, BLD may change the appearance of the base image (a, c, e), the injected regions (d), and include artifacts (b, e, f).  

\begin{figure*}[h]
  \includegraphics[width=\linewidth]{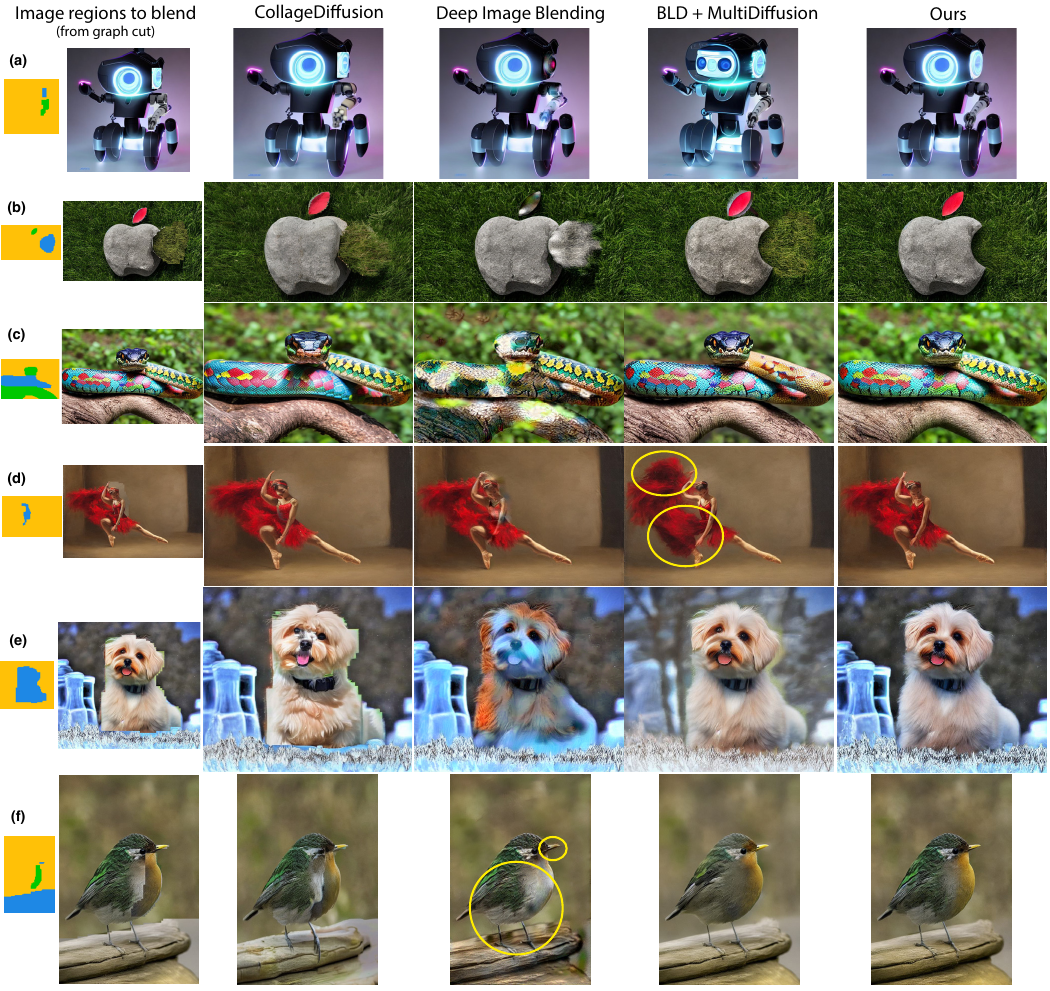}
  \caption{\textbf{Qualitative Comparison (Baselines)}. Leftmost column: Image regions to blend (output of graph-cut optimization). Graph-cut in diffusion feature space is visualized on the left, and the image-space composite of that graph-cut is visualized on the right. CollageDiffusion \cite{sarukkai2024collage} may struggle to preserve local appearance (b, c, d) or blend regions harmoniously (a, b, e, f). Deep Image Blending \cite{zhang2020deep} may also fail to preserve local appearances (a) or show artifacts in the blended regions (b-f). Blended Latent Diffusion + MultiDiffusion \cite{avrahami2023blended, bar2023multidiffusion} may show artifacts at the seams, where the noise of overlapping regions are blended together (b, e), or fail to preserve local appearances (a, c, d, f).}
  \label{fig:supp_qualitative}
\end{figure*}

\myparagraph{Blended Latent Diffusion + MultiDiffusion \cite{avrahami2023blended, bar2023multidiffusion}.} We experimented with a modified version of Blended Latent Diffusion, where we fused the noises of different image regions with greater overlap, inspired by MultiDiffusion. Specifically, we dilate the mask of each region (in feature space) with a $3\times3$ kernel and average the noise within the overlapped regions. As shown in Figure \ref{fig:supp_qualitative}, this leads to artifacts near the seams (site of overlap) or changes in local appearances.

\myparagraph{MasaCtrl with ControlNet \cite{cao2023masactrl}.} We adapt their mask-guided framework by designating the base image as the target image and the other images within the stack as source images. We extend their framework to use multiple foreground masks (one for each image beyond the base image). 
As shown in Figure~\ref{fig:supp_qualitative_full}, this may change the appearance of the base image (a, c, e), selected regions (d, e, f), or show other blending artifacts (b).

\myparagraph{CollageDiffusion \cite{sarukkai2024collage}.} We followed their released demo for generating baseline results. Their method allows users to tweak parameters per image layer and scene, such as added noise level and cross-attention modulation, and then outputs a composite. We created the input image layers by resizing the masks from feature-space graph-cut to image space and applying them to each image. Reducing the added noise to zero preserves all local regions but with little harmonization (similar to copy-paste). To minimize changes, we used a set of parameters that preserves the background region (base image) but allows the foreground injected regions to change. We set noise levels to $0.05$ for the base image and $0.4$ for the non-base images. We set noise blur to $30$ to smooth the seams. To ensure spatial fidelity of the image regions, we set the cross-attention modulation of non-base images to $0.5$, following their released examples. We also use their textual inversion feature and learn a special embedding for each image layer. If there are multiple image layers (regions) that correspond to the same word (e.g., multiple robot parts correspond to ``robot'' in the prompt ``A robot from the future''), we choose the one that most closely matches the word description.

\myparagraph{Deep Image Blending \cite{zhang2020deep}.} We ran their released code to generate the baseline results. Their method takes in two images to composite (a source and target image) and a mask of the source image. To create the mask for each image, we resized the corresponding graph-cut mask from feature to image space. For composites of more than two input images, we iteratively ran Deep Image Blending on pairs of images to build toward the final composite. One input constraint is that their method takes square images as input, so we pre-process our image stack by resizing them to square images and resizing the output back to the original aspect ratios.

\myparagraph{GP-GAN \cite{wu2019gp}.} We ran their released code with their pre-trained model to generate the results. GP-GAN takes in two images to composite (a source and destination image), along with a mask of the source image to insert into the destination image. For composites of more than two images, we iteratively ran GP-GAN on pairs of images to build toward the final composite. We created the input masks by resizing the diffusion-space graph-cut results to the original resolution of each image. 

\myparagraph{Cross-Domain Compositing \cite{hachnochi2023crossdomain}.} Cross-Domain Compositing uses pre-trained Stable Diffusion models to compose images from different domains. We ran their released code and followed their examples for object immersion. Their method takes in a composite image and a foreground mask and outputs a harmonized image. Users can control the degree of local fidelity for foreground and background regions separately, at the expense of harmonization. Due to this trade-off, their method generally cannot preserve local appearances while also blending them together harmoniously.  We created the input composite images by resizing the graph-cut masks from feature space to image space and compositing in image space. We treat the base image as the background, so the foreground mask covers regions from non-base images in the composite. To minimize changes, we applied the low-pass filter on the foreground region ($N_{in} = 2, N_{out} = 1$). To reduce artifacts at the seams, we allowed the background region to change slightly in addition to the foreground region ($T_{in} = 0.5, T_{out} = 0.9$).

\myparagraph{PCT-Net \cite{pctnet}.} 
Given a composite image and a foreground mask, PCT-Net outputs a harmonized image by applying a spatially varying, per-pixel color transformation to the masked region. We ran their released code with their pre-trained model to generate the results. We created the input composite images by resizing the feature-space graph-cut masks into image space and compositing in pixel space. We treat the base image as the background, so the foreground mask covers regions from non-base images in the composite. To smooth seams between different regions, we dilated the foreground masks  with a $17\times17$ kernel to increase overlap.

\section{User Survey Details}
\label{sec:app_user_survey}

\begin{figure}[h]
\centering
  \includegraphics[width=1.0\linewidth]{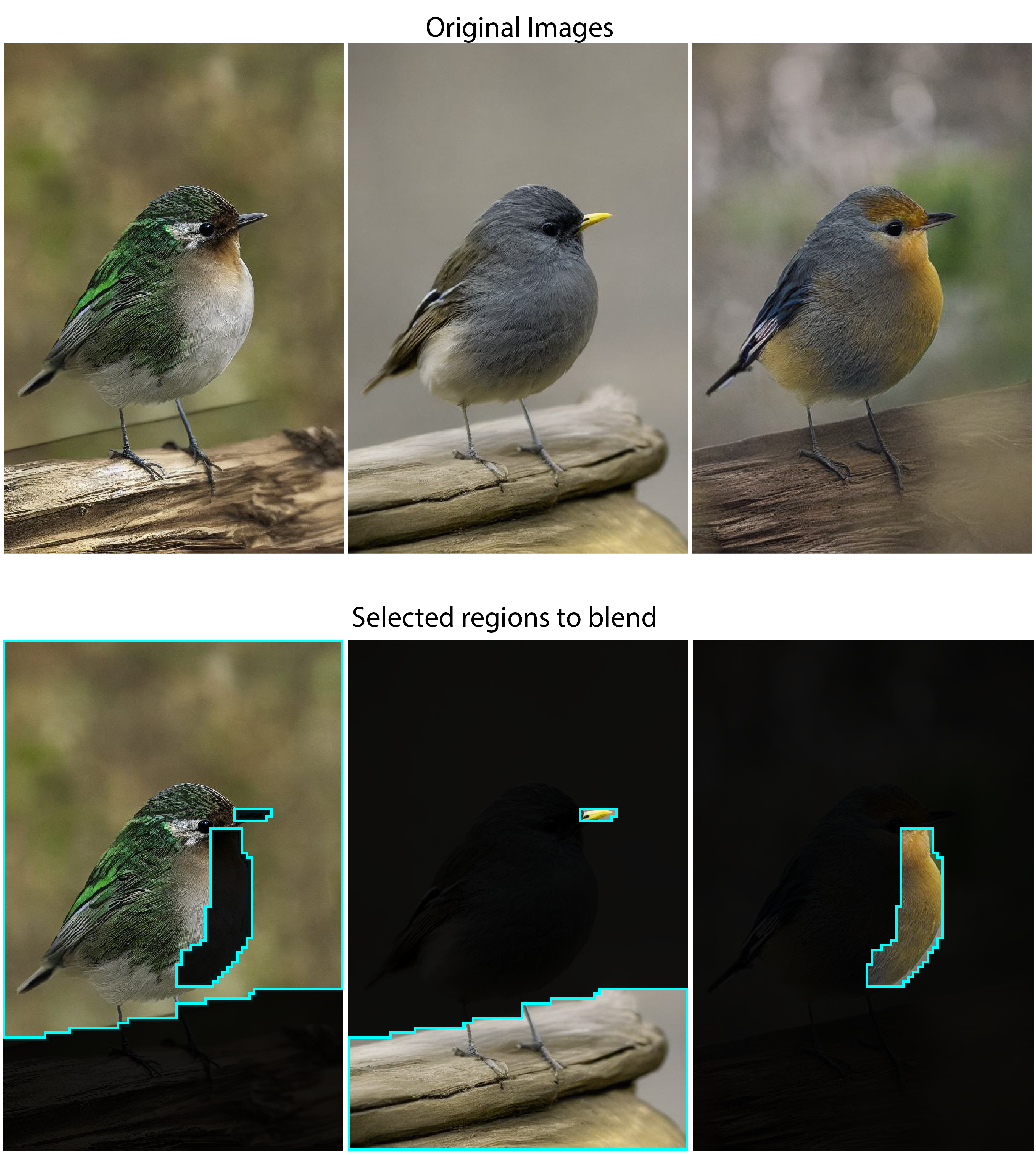}
  \caption{\textbf{User Survey (Blending)}. We show users the input images and selected regions to blend, and then ask them pick the best result from four images: one generated by our method, and the others generated by Interactive Digital Photomontage, Blended Latent Diffusion, and MasaCtrl+ControlNet.}
  \label{fig:user_survey_example}
\end{figure}

We launched two anonymous user surveys, in order to compare our method with our three most competitive baselines (Interactive Digital Photomontage \cite{agarwala2004interactive}, Blended Latent Diffusion \cite{avrahami2023blended}, MasaCtrl+ControlNet \cite{cao2023masactrl}). Across $12$ scenes (Figures \ref{fig:app_appearance1}a-f, \ref{fig:app_shapes}a-e, \ref{fig:app_fairies}), we created results using our method and the three baselines. For each scene, we showed the four results side-by-side and asked users to select the best result. The order of the four images is randomized for each scene, and the order of scenes is also randomized in each survey.

In the first survey, we asked ``Which of the following images appear most realistic to you?'' In the second survey, we showed the input images and asked users to pick the image that was the best at blending them. Specifically, we showed the full input images as well as the graph-cut regions (to be blended) on each input image. To help users focus on the individual image regions, we darken out other parts of the input image (Figure \ref{fig:user_survey_example}). Then, we asked ``Which of the following images is BEST at blending all the selected image regions shown on top?''

Users are asked to complete the first survey (realism) before the second survey (blending) to avoid bias. We received $27$ completed surveys for the realism one (a total of $324$ responses across all scenes) and $20$ completed surveys for the blending one (a total of $240$ responses across all scenes).

\section{Result Details}
\label{sec:app_result_details}
\begin{figure*}[h]
\centering
  \includegraphics[width=\linewidth]{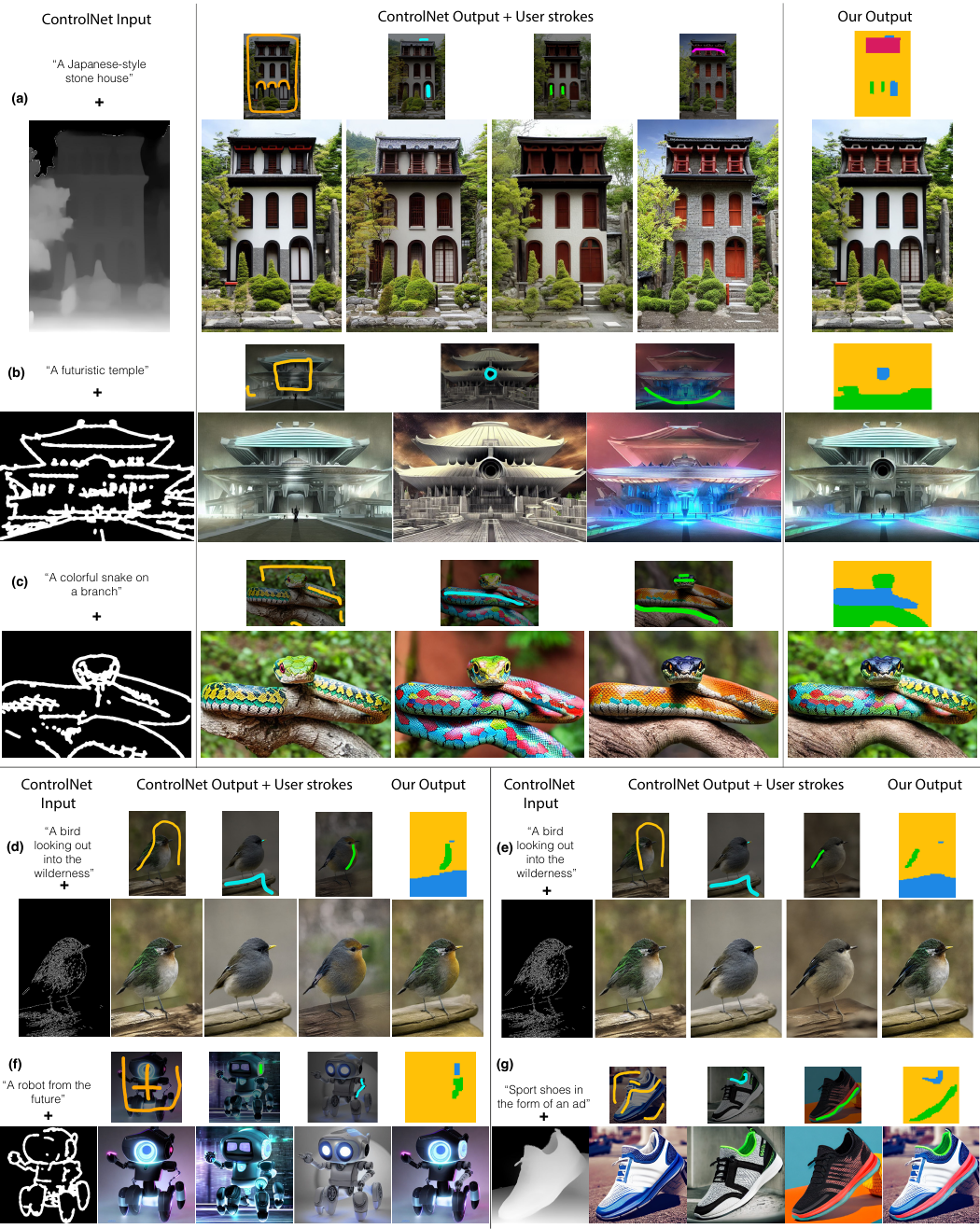}
  \caption{\textbf{Results: Appearance Mixing}. 
  Examples of using Generative Photomontage for creative design. (a, b) Users can combine different architectural elements to form new architectural designs. (c, d, e) The user combines different vibrant colors of snakes or birds to explore new looks. (f, g) The user combines different parts of a futuristic robot and shoes to form their favorite look.}
  \label{fig:app_appearance1}
\end{figure*}

\begin{figure*}[t]
\centering
  \includegraphics[width=\linewidth]{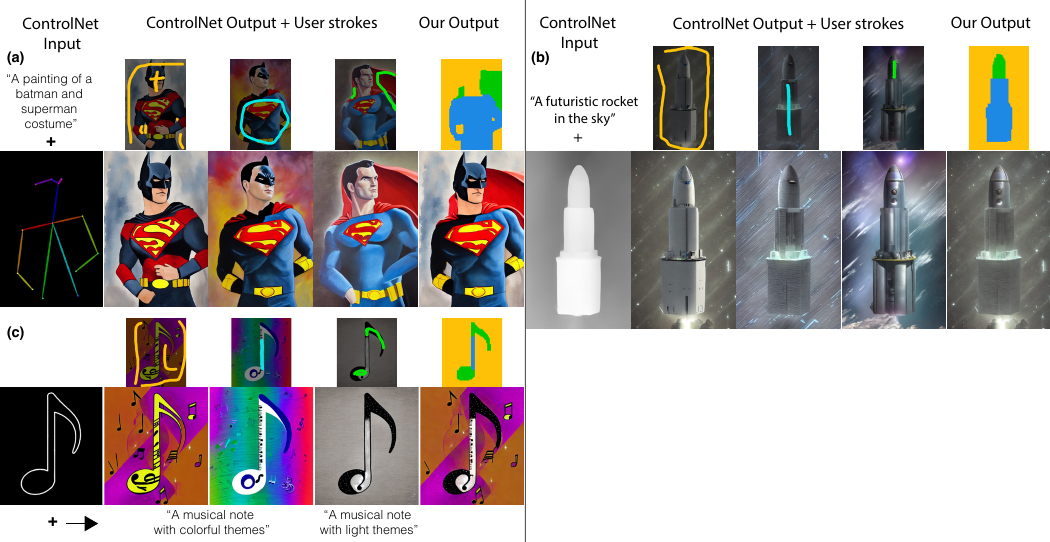}
  \caption{\textbf{Results: Appearance Mixing (cont.)}. More examples of using our method for novel artistic and graphic designs. Users can (a) mix and combine iconic features of superhero costumes to form a new one, (b) mix different visual features when exploring artistic designs of a rocket, and (c) select and combine desired elements when designing a poster.}
  \label{fig:app_appearance2}
  
\end{figure*}

\begin{figure*}[h]
\centering
  \includegraphics[width=\linewidth]{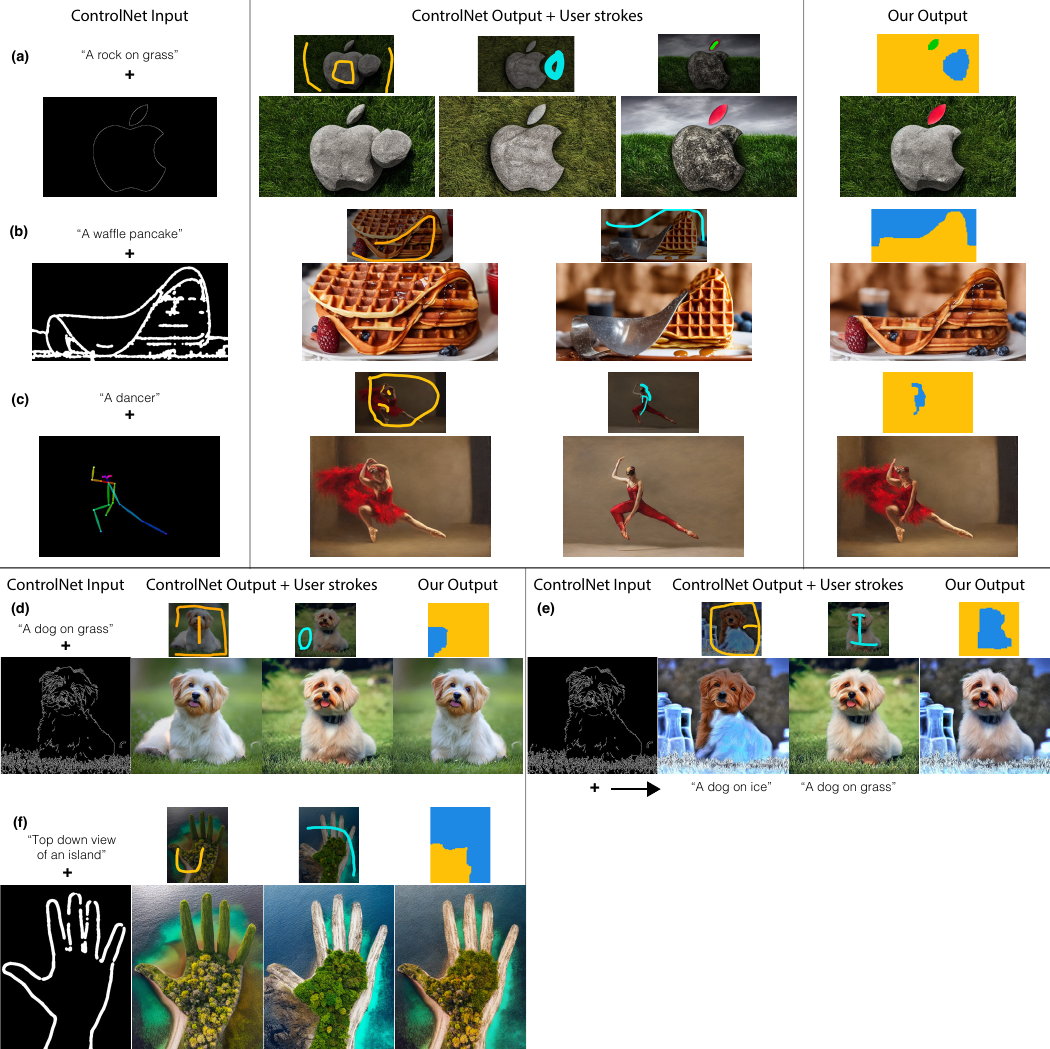}
  \caption{\textbf{Results: Shape and Artifacts Correction}. Our method can fix incorrect shapes and artifacts from ControlNet's outputs, which often occur for uncommon input shapes. For example, (a) users can remove the extra rock at the Apple bite with a patch of grass from the second image, (b) correct the shape and contour of the first waffle by selecting the background region of the second image, and (f) refine the hand-shaped island in the second image with patches from the first image. Users can also (c) correct the dancer's pose, (d) remove the extra leg of the dog, and (e) and replace the first dog with artifacts with the second dog.}
  \label{fig:app_shapes}
\end{figure*}

We used various Stable Diffusion pre-trained models (v1.5) to generate results in this paper. Specifically, we used the canny edge model (Figures \ref{fig:app_appearance1}d-e, \ref{fig:app_shapes}a,d,e, \ref{fig:app_fairies}), the scribble model (Figures \ref{fig:app_appearance1}b,c,f, \ref{fig:app_shapes}b, f), the depth map model (Figure \ref{fig:app_appearance1}a, g, \ref{fig:app_appearance2}b, \ref{fig:lamp}), the Openpose model (Figures \ref{fig:app_appearance2}a, \ref{fig:app_shapes}c), and the HED model (Figures \ref{fig:app_appearance2}c, \ref{fig:red}, \ref{fig:eye}, \ref{fig:limitations}). 

The input conditions (e.g., edge map, sketches) are manually created or derived from images released by ControlNet \cite{zhang2023adding} or found on the web \cite{snake_source, dancer_pose, temple_source, waffle_source, red_source, fairy_source, musical_source, hand_source, eye_source, lipstick_source, lamp_source}.

\section{Limitations}
\label{sec:app_limitations}

\begin{figure*}
  \vspace{-5pt}
  \centering
  \includegraphics[width=0.5\linewidth]{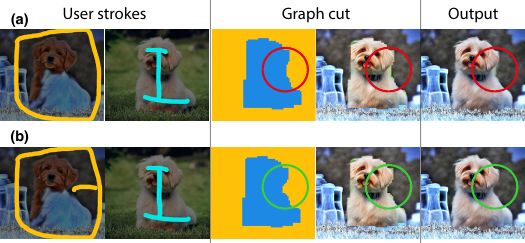}
  \caption{\textbf{Limitation}. Our graph cut optimization prefers smaller seam circumferences due to lower pairwise costs (Equation \ref{eq:pairwise}). For objects with curved outlines, users may need to refine boundaries with additional strokes. %
  (a) Our graph cut over-segmented the dog at its neck region because a vertical seam (circled in red) has a lower circumference (and lower cost) than a curved one. (b) To refine the boundary, users can add an additional stroke in the background, which aligns the boundary to the dog's neck (circled in green).}
  \label{fig:limitations_gc}
  \vspace{-10pt}
\end{figure*}

While we have shown our method's versatility in various applications, we also observe several limitations. First, our current graph cut parameters are empirically chosen to encourage congruous regions, which penalizes seam circumference. While this works well for many cases, if the target object has a curvy outline, it may require additional user strokes to obtain a finer boundary (Figure \ref{fig:limitations_gc}). Since graph cut is solved in near real-time (${\sim}1$s), users can quickly check the graph-cut result and iterate as needed. 

\begin{figure*}
  \includegraphics[width=\textwidth]{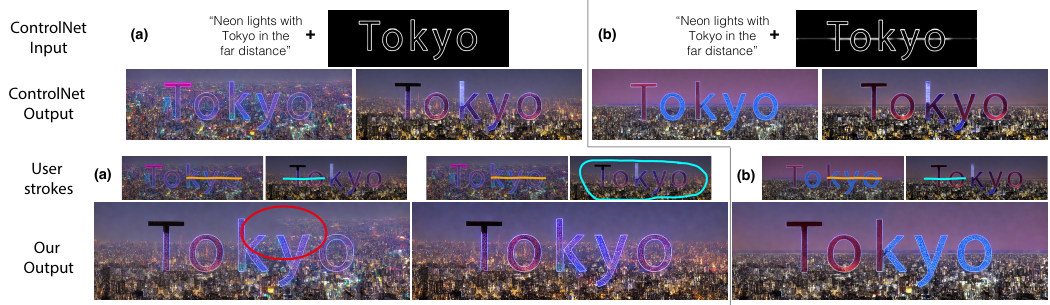}
  \caption{\textbf{Limitation}. Our method assumes some spatial consistency among images in the stack. In cases where the images differ significantly in scene structure, our method may produce semantically incorrect outputs. (a) Two images have different horizons in the background. Naively combining two halves of the images leads to an inconsistent horizon (bottom left, circled red). Users can manually designate a consistent horizon by selecting the background of the second image (bottom, middle). (b) Alternatively, users can add a horizon in the input sketch to ControlNet to make it consistent across both images.}
  \label{fig:limitations}
  \vspace{-10pt}
\end{figure*}

Second, our method assumes some spatial consistency among images in the stack. If the images differ significantly in scene structure, it will rely more on the user to select proper regions to form a valid scene (Figure \ref{fig:limitations}a). Alternatively, users can increase spatial consistency by adding more spatial structure to the input control (Figure \ref{fig:limitations}b). Future work can investigate ways to relax spatial consistency constraints and automatically account for dramatic scene structure changes during segmentation and blending.

\section{SDXL Results}
\begin{figure*}[h]
\centering
  \includegraphics[width=\linewidth]{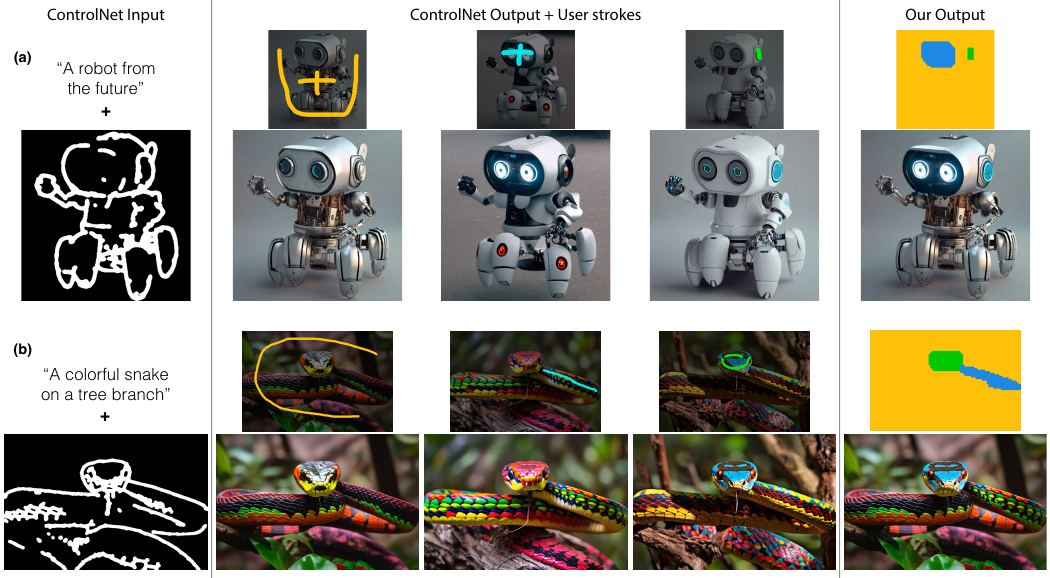}
  \caption{\textbf{Generative Photomontage with SDXL}. 
  Examples of using Generative Photomontage on SDXL.}
  \label{fig:sdxl_results}
\end{figure*}

Our initial tests show that our method works on SDXL as well. Please see Figure \ref{fig:sdxl_results} for example results. 

For SDXL, we kept all hyperparameters the same, except for the graph cut parameter $\sigma$, which we increased to $25$. We found that using the same $\sigma$ as in SD1.5 results in more cuts and less coherent segmentations in SDXL. This may be due to SDXL's larger feature map size and differences in how feature variations correspond to semantic boundaries.


\section{Changelog}

\textbf{v1}: Original draft.

\noindent \textbf{v2}: Fixed typos.

\noindent \textbf{v3}: Updated with CVPR 2025 camera ready version. Updated user survey with more responses.

\end{document}